\documentclass{article}

    \PassOptionsToPackage{numbers, compress}{natbib}

\usepackage{amsmath, amssymb}
\usepackage{bm}
\usepackage{empheq}
\usepackage{algorithm}
\usepackage{algorithmic}
\usepackage{wrapfig}

 \usepackage[preprint]{neurips_2025}

\usepackage[utf8]{inputenc} %
\usepackage[T1]{fontenc}    %
\usepackage{hyperref}       %
\usepackage{url}            %
\usepackage{booktabs}       %
\usepackage{amsfonts}       %
\usepackage{nicefrac}       %
\usepackage{microtype}      %
\usepackage{xcolor}         %

\usepackage{graphicx}
\usepackage{subcaption}
\usepackage[font=small]{caption}

\title{Flow Matching Policy Gradients}

\author{
{David McAllister$^{1}$\thanks{Equal contribution.} ~ Songwei Ge$^{1*}$ ~ Brent Yi{$^{1*}$} ~ Chung Min Kim{$^{1}$}} \\ \textbf{Ethan Weber{$^{1}$ ~ Hongsuk Choi{$^{1}$} ~ Haiwen Feng{$^{1,2}$} ~ Angjoo Kanazawa{$^{1}$}}} \\
\normalsize
$^{1}$\	UC Berkeley  
$^{2}$\ Max Planck Institute for Intelligent Systems 
}

\begin{document}

 \maketitle

\begin{abstract}

Flow-based generative models, including diffusion models, excel at modeling continuous distributions in high-dimensional spaces. In this work, we introduce Flow Policy Optimization (FPO), a simple on-policy reinforcement learning algorithm that brings flow matching into the policy gradient framework. FPO casts policy optimization as maximizing an advantage-weighted ratio computed from the conditional flow matching loss, in a manner compatible with the popular PPO-clip framework. It sidesteps the need for exact likelihood computation while preserving the generative capabilities of flow-based models. Unlike prior approaches for diffusion-based reinforcement learning that bind training to a specific sampling method, FPO is agnostic to the choice of diffusion or flow integration at both training and inference time. %
We show that FPO can train diffusion-style policies from scratch in a variety of continuous control tasks.
We find that flow-based models can capture multimodal action distributions and achieve higher performance than Gaussian policies, particularly in under-conditioned settings. For an overview of FPO's key ideas, see our accompanying blog post: \href{https://flowreinforce.github.io}{flowreinforce.github.io}

\end{abstract}

\section{Introduction}

Flow-based generative models---particularly diffusion models---have emerged as powerful tools for generative modeling across the domains of images~\cite{ramesh2022hierarchical,saharia2022photorealistic,ho2022imagen}, videos~\cite{videoworldsimulators2024,polyak2024movie,veo2}, speech~\cite{liu2023audioldmtexttoaudiogenerationlatent}, audio~\citep{kong2021diffwaveversatilediffusionmodel}, robotics~\cite{chi2024diffusionpolicy}, and molecular dynamics~\cite{raja2025action}. In parallel, reinforcement learning (RL) has proven to be effective for optimizing neural networks with non-differentiable objectives, and is widely used as a post-training strategy for aligning foundation models with task-specific goals~\citep{chu2025sft, liu2024deepseek}.

In this work, we introduce Flow Policy Optimization (FPO), a policy gradient algorithm for optimizing flow-based generative models. FPO reframes policy optimization as maximizing an advantage-weighted ratio computed from the conditional flow matching (CFM) objective~\cite{lipman2023flowmatchinggenerativemodeling}. Intuitively, FPO shapes probability flow to transform Gaussian noise into high-reward actions by reinforcing its experience using flow matching.
The method is simple to implement and can be readily integrated into standard techniques for stochastic policy optimization. %
We use a PPO-inspired surrogate objective for our experiments, which trains stably and serves as a drop-in replacement for Gaussian policies.

FPO offers several key advantages. It sidesteps the complex likelihood calculations typically associated with flow-based models, instead using the flow matching loss as a surrogate for log-likelihood in the policy gradient. This aligns the objective directly with increasing the evidence lower bound of high-reward actions. Unlike previous methods that reframe the denoising process as an MDP, binding the training to specific sampling methods and extending the credit-assignment horizon, FPO treats the sampling procedure as a black box during rollouts. This distinction allows for flexible integration with any sampling approach---whether deterministic or stochastic, first- or higher-order, and with any number of integration steps during training or inference.

We theoretically analyze FPO’s correctness and empirically validate its performance across a diverse set of tasks. These include a GridWorld environment, 10 continuous control tasks from MuJoCo Playground~\cite{zakka2025mujoco}, and high-dimensional humanoid control---all trained from scratch. FPO demonstrates robustness across tasks, enabling effective training of flow-based policies in high-dimensional domains. We probe flow policies learned in the toy GridWorld environment and find that on states with multiple possible optimal actions, it learns multimodal action distributions, unlike Gaussian policies.
On humanoid control tasks, we show that the expressivity of flow matching enables single-stage training of under-conditioned control policies, where only root-level commands are provided. In contrast, standard Gaussian policies struggle to learn viable walking behaviors in such cases. This highlights the practical benefits of the more powerful distribution modeling enabled by FPO. Finally, we discuss limitations and future work.

\section{Related Work}
\label{related}

\noindent\textbf{Policy Gradients.}
We study on-policy reinforcement learning, where a parameterized policy is optimized to maximize cumulative reward in a provided environment.
This is commonly solved with policy gradient techniques, which bypass the need for differentiable environment rewards by weighting action log-probabilities with observed rewards or advantages~\citep{sutton1999policygradient,williams1992simple,kakade2002naturalpg,peters2008nac,schulman2015trust,schulman2017proximal,mnih2016a3c,wang2016acer,shao2024deepseekmath}.
Policy gradient methods are central in learning policies for general continuous control tasks~\citep{duan2016benchmarking,huang2024open}, robot locomotion~\citep{rudin2022learning,schwarke2023curiosity,mittal2024symmetry,videomimic} and manipulation~\citep{akkaya2019solving,chen2021system,qi2023general,qi2025simple}.
They have also been adopted increasingly for searching through and refining prior distributions in pretrained generative models. This has proven effective for alignment with human preferences~\cite{ouyang2022training, christiano2023deepreinforcementlearninghuman} and improving reasoning using verifiable rewards~\cite{deepseekai2025deepseekr1incentivizingreasoningcapability, mistralai2025magistral}.

In this work, we propose a simple algorithm for training flow-based generative policies, such as diffusion models, under the policy gradient framework. By leveraging recent insights from flow matching~\citep{lipman2023flowmatchinggenerativemodeling}, we train policies that can represent richer distributions than the diagonal Gaussians that are most frequently used for reinforcement learning for continuous control~\citep{rudin2022learning,schwarke2023curiosity,mittal2024symmetry,videomimic,qi2023general,qi2025simple}, while remaining compatible with standard actor-critic training techniques.

\noindent\textbf{Diffusion Models.}
Diffusion models are powerful tools for modeling complex continuous distributions and have achieved remarkable success across a wide range of domains. These models have become the predominant approach for generating images~\cite{ho2020denoising,song2022denoisingdiffusionimplicitmodels,rombach2022highresolutionimagesynthesislatent,song2020generativemodelingestimatinggradients}, videos~\cite{ho2022videodiffusionmodels,singer2022makeavideotexttovideogenerationtextvideo,ho2022imagenvideohighdefinition,videoworldsimulators2024}, audio~\cite{liu2023audioldmtexttoaudiogenerationlatent,popov2021gradttsdiffusionprobabilisticmodel,chen2021wavegrad2iterativerefinement,kong2021diffwaveversatilediffusionmodel}, and more recently, robot actions~\cite{chi2024diffusionpolicy,black2024pi0visionlanguageactionflowmodel,nvidia2025gr00tn1openfoundation}. In these applications, diffusion models aim to sample from a data distribution of interest, whether scraped from the internet or collected through human teleoperation.

Flow matching~\citep{lipman2023flowmatchinggenerativemodeling} simplifies and generalizes the diffusion model framework. It learns a vector field that transports samples from a tractable prior distribution to the target data distribution. The conditional flow matching (CFM) objective trains the model to denoise data that has been perturbed with Gaussian noise. Given data $x$ and noise $\epsilon \in \mathcal{N}(0, I)$, the CFM objective can be expressed as:
\begin{align}
\mathcal{L}_{\text{CFM}, \theta} = \mathbb{E}_{\tau, q(x), p_\tau(x_\tau \mid x)} \left\| \hat{v}_{\theta}(x_\tau, \tau) - u(x_\tau, \tau \mid x) \right\|_2^2,
\label{eq:conditional_flow_matching}
\end{align}
where $x_\tau = \alpha_\tau x + \sigma_\tau \epsilon$ represents the partially noised sample at flow step $\tau$, an interpolation of noise and data with a schedule defined by hyperparameters $\alpha_\tau$ and $\sigma_\tau$. $\hat{v}_\theta(x_\tau, \tau)$ is the model's estimate of the velocity to the original data, and $u(x_\tau, \tau \mid x)$ is the conditional flow $x - \epsilon$. The model can also estimate the denoised sample $x$ or noise component $
\epsilon$ as the optimization target instead of velocity. The learned velocity field is a continuous mapping that transports samples from a simple, tractable distribution (e.g. Gaussian noise) to the training data distribution through ODE integration.

Optimizing likelihoods directly through flow models is possible, but requires divergence estimation~\cite{skreta2025superpositiondiffusionmodelsusing} and is computationally prohibitive. Instead, flow matching optimizes variational lower bounds of the likelihood with the simple denoising loss above. In this work, we leverage flow matching directly within the policy gradient formulation. This approach trains diffusion models from rewards without prohibitively expensive likelihood computations.

\noindent\textbf{Diffusion Policies.}
Diffusion-based policies have shown promising results in robotics and decision-making applications~\citep{chi2023diffusion, ajay2022conditional, black2024pi0visionlanguageactionflowmodel}. Most existing approaches train these models via behavior cloning~\citep{janner2022planning,chi2024diffusionpolicy}, where the policy is supervised to imitate expert trajectories without using reward feedback. Motivated by the strong generative capabilities of diffusion and flow-based models, several works have explored using reinforcement learning to fine-tune diffusion models, particularly in domains like text-to-image generation~\citep{lee2023aligning, black2023training,liu2025flow}.

Recent work by Psenka et al.~\citep{psenka2023learning} explores off-policy training of diffusion policies via Q-score matching. While off-policy reinforcement learning continues to make progress~\cite{seo2025fasttd3simplefastcapable,fujimoto2018addressingfunctionapproximationerror}, on-policy methods dominate practical applications today. Methods like DDPO~\citep{black2023training}, DPPO~\citep{ren2024diffusion}, and Flow-GRPO~\citep{liu2025flow} adopt on-policy policy gradient methods by treating initial noise values as observations from the environment, framing the denoising process as a Markov decision process, and training each step as a Gaussian policy using PPO. Our approach differs by directly integrating the conditional flow matching (CFM) objective into a PPO-like framework, maintaining the structure of the standard diffusion forward and reverse processes. %
Since FPO integrates flow matching as its fundamental primitive, it is agnostic to the choice of sampling method during both training and inference, just like flow matching for behavior cloning.

\section{Flow Matching Policy Gradients}
\label{Methods}

\subsection{Policy Gradients and PPO}

The goal of reinforcement learning is to learn a policy $\pi_\theta$ that maximizes expected return in a provided environment.
At each iteration of online reinforcement learning, the policy is rolled out to collect batches of observation, action, and reward tuples $(o_t, a_t, r_t)$ for each environment timestep $t$.
These rollouts can used in the policy gradient objective~\cite{sutton1999policygradient} to increase likelihood of actions that result in higher rewards:
\begin{align}
    \max_\theta \ \mathbb{E}_{a_t \sim \pi_\theta(a_t \mid o_t)} \left[
        \log \pi_\theta(a_t \mid o_t)
        \hat{A}_t
    \right] ,
    \label{eq:pg_objective}
\end{align}
where $\hat{A}_t$ is an advantage estimated from the rollout's rewards $r_t$ and a learned value function~\citep{schulman2015high}.

The vanilla policy gradient is valid only locally around the current policy parameters. Large updates can lead to policy collapse or unstable learning. To address this, PPO~\cite{schulman2017proximal} incorporates a trust region by clipping the likelihood ratio: %
\providecommand{\pporatio}[0]{\frac{\pi_\theta(a_t \mid o_t)}{\pi_\text{old}(a_t \mid o_t)}}
\providecommand{\ppoeps}[0]{\ensuremath{\varepsilon^\text{clip}}}
\begin{align}
\max_\theta \ \mathbb{E}_{a_t \sim \pi_{\theta_\text{old}}(a_t\mid o_t)} \left[ \min \left( r(\theta) \hat{A}_t, \, \text{clip}(r(\theta), 1 - \ppoeps{}, 1 + \ppoeps{}) \hat{A}_t \right) \right],
\label{eq:ppo}
\end{align}
where $\ppoeps$ is a tunable threshold and $r(\theta)$ is the ratio between current and old action likelihoods:
\begin{align}
    r(\theta) =\pporatio.
    \label{eq:ppo_ratio_definition}
\end{align}

PPO is popular choice for on-policy reinforcement learning because of its stability, simplicity, and performance.
Like the standard policy gradient, however, it requires exact likelihoods for sampled actions.
These quantities are tractable for simple Gaussian or categorical action spaces, but computationally prohibitive to estimate for flow matching and diffusion models.

\subsection{Flow Policy Optimization}
\label{sec:flow_policy_optimization}

\providecommand{\att}[0]{\ensuremath{a_t^\tau}}

We introduce Flow Policy Optimization (FPO), an online reinforcement learning algorithm for policies represented as flow models $\hat{v}_\theta$. 
There are two key differences in practice from Gaussian PPO. During rollouts, a flow model transforms random noise into actions via a sequence of learned transformations~\citep{lipman2023flowmatchinggenerativemodeling}, enabling much more expressive policies than those used in standard PPO. Also, to update the policy, the Gaussian likelihoods are replaced with a transformed flow matching loss. %

Instead of updating exact likelihoods, 
we propose a proxy $\hat{r}^\text{FPO}$ for the log-likelihood ratio.
FPO's overall objective is the same as Equation~\ref{eq:ppo}, but with the ratio substituted:
\providecommand{\ppoeps}[0]{\ensuremath{\varepsilon^\text{clip}}}
\begin{align}
    \max_\theta \ \mathbb{E}_{a_t \sim \pi_\theta(a_t\mid o_t)} \left[ \min \left( \hat{r}^{\text{FPO}}(\theta) \hat{A}_t, \, \text{clip}(\hat{r}^{\text{FPO}}(\theta), 1 - \ppoeps{}, 1 + \ppoeps{}) \hat{A}_t \right) \right].
\label{eq:fpo}
\end{align}
Intuitively, FPO's goal is to steer the policy's probability flow toward high-return behavior. Instead of computing likelihoods, we construct a simple ratio estimate using standard flow matching losses:
\begin{align}
    \hat{r}^\text{FPO}(\theta) = \exp( \hat{\mathcal{L}}_{\text{CFM},{\theta_\text{old}}}(a_t; o_t) - \hat{\mathcal{L}}_{\text{CFM},\theta}(a_t; o_t)),
    \label{eq:fpo_ratio_in_practice}
\end{align}
which, as we will discuss, can be derived from optimizing the evidence lower bound.

For a given action and observation pair, $\hat{\mathcal{L}}_{\text{CFM},\theta}(a_t; o_t)$ is an estimate of the per-sample conditional flow matching loss $\mathcal{L}_{\text{CFM},\theta}(a_t; o_t)$:
\begin{align}
    \hat{\mathcal{L}}_{\text{CFM}, \theta}(a_t ; o_t) &= \frac{1}{N_\text{mc}} \sum_i^{N_\text{mc}} \ell_\theta(\tau_i, \epsilon_i)
    \label{eq:cfm_loss_monte_carlo_estimator}\\
    \ell_\theta(\tau_i, \epsilon_i) &= \lvert\lvert 
        \hat{v}_{\theta}(a_t^{\tau_i}, \tau_i; o_t) - (a_t - \epsilon_i)
    \rvert\rvert_2^2\\
    a_t^{\tau_i} &= \alpha_{\tau_i} a_t + \sigma_{\tau_i} \epsilon_i,
\end{align}
where we denote flow timesteps with $\tau$ and environment timesteps with $t$.
We include both timesteps in $\att$, which represents an action at rollout time $t$ with noise level $\tau$ following Equation~\ref{eq:conditional_flow_matching}.
We use the same $\epsilon_i \sim N(0, I)$ and $\tau_i \in [0, 1]$ samples between $\hat{\mathcal{L}}_{\text{CFM}, \theta_\text{old}}$ and $\hat{\mathcal{L}}_{\text{CFM}, \theta}$.

\textbf{Properties.} FPO's ratio estimate in Equation~\ref{eq:fpo_ratio_in_practice} serves as a drop-in replacement for the PPO likelihood ratio.
FPO therefore inherits compatibility with advantage estimation methods like GAE~\cite{schulman2015high} and GRPO~\cite{shao2024deepseekmath}.
Without loss of generality, it is also compatible with flow and diffusion implementations based on estimating noise $\epsilon$~\citep{ho2020denoising} or clean action $a_t$~\citep{ramesh2022hierarchical}, which can be reweighted for mathematical equivalence to $\mathcal{L}_{\theta,\text{CFM}}$~\citep{karras2022elucidating,lipman2023flowmatchinggenerativemodeling}.
We leverage this property in our FPO ratio derivation below. %

\subsection{FPO Surrogate Objective}

Exact likelihood is computationally expensive even to estimate in flow-based models.
Instead, it is common to optimize the evidence lower bound (ELBO) as a proxy for log-likelihood:
\begin{align}
    \text{ELBO}_\theta(a_t \mid o_t) = \log \pi_\theta(a_t \mid o_t) - \mathcal{D}_\theta^\text{KL} ,
    \label{eq:elbo_definition}
\end{align}
where $\mathcal{D}_\theta^\text{KL}$ is the KL gap between the ELBO and true log-likelihood and $\pi_\theta$ is the distribution captured by sampling from the flow model. Both flow matching and diffusion models optimize the ELBO using a conditional flow matching loss, a simple MSE denoising objective~\cite{kingma2023variationaldiffusionmodels,lipman2023flowmatchinggenerativemodeling}.
The FPO ratio (Equation~\ref{eq:fpo_ratio_definition}) leverages the fact that flow models can be trained via ELBO objectives. Specifically, we compute the ratio of ELBOs under the current and old policies:%
\begin{align}
    r^{\text{FPO}}(\theta)=\frac{\exp(\text{ELBO}_\theta(a_t \mid o_t))}{\exp(\text{ELBO}_{\theta_\text{old}}(a_t \mid o_t))}.
    \label{eq:fpo_ratio_definition}
\end{align}
Decomposing this ratio reveals a scaled variant of the true likelihood ratio (Equation~\ref{eq:ppo_ratio_definition}):
\begin{align}
    r^{\text{FPO}}(\theta)=\underbrace{\frac{\pi_\theta(a_t \mid o_t)}{\pi_{\theta_\text{old}}(a_t \mid o_t)}}_{\text{Likelihood}} \underbrace{\frac{\exp(\mathcal{D}_{\theta_\text{old}}^\text{KL})}{\exp(\mathcal{D}_\theta^\text{KL})}}_{\text{Inv. KL Gap}}.
\end{align}
Here, the ratio decomposes into the standard likelihood ratio and an inverse correction term involving the KL gap. Maximizing this ratio therefore increases the modeled likelihood while reducing the KL gap---both of which are beneficial for policy optimization. The former encourages the policy to favor actions with positive advantage, while the latter tightens the approximation to the true log-likelihood.

\subsection{Estimating the FPO Ratio with Flow Matching}

\providecommand{\weightedloss}[0]{\ensuremath{\mathcal{L}_\theta^w}}
\providecommand{\weightedlossmc}[0]{\ensuremath{\ell_\theta^w}}
\providecommand{\fporatiomc}[0]{\ensuremath{\hat{r}^{\text{FPO}}_\theta(\tau, \epsilon)}}

We estimate the FPO ratio using the flow matching objective directly, which follows from the relationship between the weighted denoising loss $\weightedloss{}$ and the ELBO established by Kingma and Gao~\cite{kingma2023understandingdiffusionobjectiveselbo}. $\weightedloss$ is a more general form of the flow matching and denoising diffusion loss that parameterizes the model as predicting $\hat{\epsilon}_\theta$, an estimate of the true noise $\epsilon$ present in the model input.

The weighted denoising loss \weightedloss\; for a clean action $a_t$ takes the form:
\begin{align}
\weightedloss(a_t) = \frac{1}{2} \mathbb{E}_{\tau \sim \mathcal{U}(0,1), \epsilon \sim \mathcal{N}(0,I)} \left[ w(\lambda_\tau) \cdot \left(-\frac{d\lambda}{d\tau}\right) \cdot \|\hat{\epsilon}_\theta(\att; \lambda_\tau) - \epsilon\|^2_2 \right],
\end{align}
where $w$ is a choice of weighting and $\lambda_\tau$ represents the log-SNR at noise level $\tau$.
We estimate this value with Monte Carlo draws of timestep $\tau$ and noise $\epsilon$:
\begin{align}
    \ell_\theta^w(\tau, \epsilon) = \frac{1}{2} w(\lambda_\tau) \cdot \left(-\frac{d\lambda}{d\tau}\right) \cdot \|\hat{\epsilon}_\theta(\att; \lambda_\tau) - \epsilon\|^2_2.
\end{align}
The choice of weighting $w$ incorporates the conditional flow matching loss and standard diffusion loss as specific cases of a more general family $\weightedloss(a_t)$.

We focus here on the constant weight case $w(\lambda_\tau) = 1$ (diffusion schedule), which yields the simplest theoretical connection. Similar results hold for many popular schedules, including optimal transport and variance preserving schedules~\cite{lipman2023flowmatchinggenerativemodeling}. Please see the supplementary material for details.

For the diffusion schedule, \cite{kingma2023understandingdiffusionobjectiveselbo} proves that:
\begin{align}
\weightedloss(a_t) = -\text{ELBO}_\theta(a_t) + c,
\end{align}
where $c$ is a constant w.r.t $\theta$. Geometrically, minimizing $\weightedloss(a_t)$
points the flow more toward $a_t$. Minimizing $\weightedloss$ also maximizes the ELBO (Eq.~\ref{eq:elbo_definition}) and thus the likelihood of $a_t$, so flowing toward a specific action makes it more likely. This intuition aligns naturally with the policy gradient objective: we want to increase the probability of high-advantage actions. By redirecting flow toward such actions (i.e., minimizing their diffusion loss), we make them more likely under the learned policy.

Using this relationship, we express the FPO ratio (Eq.~\ref{eq:fpo_ratio_definition}) in terms of the flow matching objective:
\begin{align}
r^{\text{FPO}}_\theta = \frac{\exp(\text{ELBO}_\theta(a_t | o_t))}{\exp(\text{ELBO}_{\theta_{\text{old}}}(a_t | o_t))} = \exp(\mathcal{L}^w_{\theta_{\text{old}}}(a_t) - \mathcal{L}^w_{\theta}(a_t)),
\label{fporatio}
\end{align}
where $\mathcal{L}^w_\theta$, as per Equation~\ref{eq:cfm_loss_monte_carlo_estimator}, can be estimated by averaging over $N_\text{mc}$ draws of ($\tau$, $\epsilon$).
We find the sample count $N_\text{mc}$ to be a useful hyperparameter for controlling learning efficiency.
This estimator recovers the exact FPO ratio in the limit, although we use only a few draws in practice.

One possible concern with smaller $N_\text{mc}$ values is bias.
A ratio estimated from only one ($\tau$, $\epsilon$) pair,
\begin{align}
\hat{r}^{\text{FPO}}_\theta(\tau, \epsilon) = \exp(\ell^w_{\theta_{\text{old}}}(\tau, \epsilon)-\ell^w_\theta(\tau, \epsilon)),
\end{align}
is in expectation only an upper-bound of the true ratio.
This can be shown by Jensen's inequality:
\begin{align}
\mathbb{E}_{\tau,\epsilon}[\hat{r}^{\text{FPO}}_\theta(\tau, \epsilon)] \geq r^{\text{FPO}}_\theta.
\end{align}
To understand the upward bias, we can use the log-derivative trick to decompose the FPO gradient:
\begin{align}
\nabla_\theta \fporatiomc=-\fporatiomc \nabla_\theta \ell^w_\theta(\tau, \epsilon).
\end{align}
Since the gradient operator commutes with expectation, the gradient term on the right side is unbiased: %
\begin{align}
\mathbb{E}_{\tau,\epsilon}[-\nabla_\theta \ell^w_\theta(\tau, \epsilon)] = -\nabla_\theta \weightedloss{}(a_t) = \nabla_\theta \text{ELBO}_\theta(a_t).
\end{align}
In other words, gradient estimates are directionally unbiased even with worst-case overestimation of ratios. %
Our experiments are consistent with this result: while additional samples help, we observe empirically in Section~\ref{sec:mujoco} that FPO can be trained to outperform Gaussian PPO even with $N_\text{mc} = 1$.

Algorithm~\ref{alg:fpo} details FPO's practical implementation using this mathematical framework.

\begin{algorithm}[t]
\caption{Flow Policy Optimization (FPO)}
\label{alg:fpo}
\begin{algorithmic}[1]
\REQUIRE Policy parameters $\theta$, value function parameters $\phi$, clip parameter $\epsilon$, MC samples $N_{\text{mc}}$
\WHILE{not converged}
    \STATE Collect trajectories using any flow model sampler and compute advantages $\hat{A}_t$
    \STATE For each action, store $N_{\text{mc}}$ timestep-noise pairs $\{(\tau_i, \epsilon_i)\}$ and compute $\ell_{\theta}(\tau_i, \epsilon_i)$
    \STATE $\theta_{\text{old}} \leftarrow \theta$
    \FOR{each optimization epoch}
        \STATE Sample mini-batch from collected trajectories
        \FOR{each state-action pair $(o_t, a_t)$ and corresponding MC samples $\{(\tau_i, \epsilon_i)\}$}
            \STATE Compute $\ell_\theta(\tau_i, \epsilon_i)$ using stored $(\tau_i, \epsilon_i)$
            \STATE $\hat{r}_\theta \leftarrow \exp\left(-\frac{1}{N_{\text{mc}}} \sum_{i=1}^{N_{\text{mc}}} (\ell_\theta(\tau_i, \epsilon_i) - \ell_{\theta_{\text{old}}}(\tau_i, \epsilon_i))\right)$ %
            \STATE $L^{\text{FPO}}(\theta) \leftarrow \min(\hat{r}_\theta \hat{A}_t, \text{clip}(\hat{r}_\theta, 1 \pm \epsilon) \hat{A}_t)$
        \ENDFOR
        \STATE $\theta \leftarrow \text{Optimizer}(\theta, \nabla_\theta \sum L^{\text{FPO}}(\theta))$
    \ENDFOR
    \STATE Update value function parameters $\phi$ like standard PPO
\ENDWHILE
\end{algorithmic}
\end{algorithm}

\subsection{Denoising MDP Comparison}

Existing algorithms~\cite{black2023training,ren2024diffusion,liu2025flow} for on-policy reinforcement learning with diffusion models reformulate the denoising process itself as a Markov Decision Process (MDP).
These approaches bypass flow model likelihoods by instead treating every step in the sampling chain as its own action, each parameterized as a Gaussian policy step.
This has a few limitations that FPO addresses.

First, denoising MDPs multiply the horizon length by the number of denoising steps (typically 10-50), which increases the difficulty of credit assignment.
Second, these MDPs do not consider the initial noise sample during likelihood computation.
Instead, these noise values are treated as observations from the environment~\cite{ren2024diffusion}---this significantly increases the dimensionality of the learning problem.
Finally, denoising MDP methods are limited to stochastic sampling procedures by construction.
Instead, since FPO employs flow matching, it inherits the flexibility of sampler choices from standard flow/diffusion models.
These include fast deterministic samplers, higher-order integration, and choosing any number of sampling steps. Perhaps most importantly, FPO is simpler to implement because it does not require a custom sampler or the notion of extra environment steps.

\section{Experiments}
\label{experiments}
We assess FPO's effectiveness by evaluating it in multiple domains. Our experiments include: (1)~an illustrative GridWorld environment using Gymnasium~\citep{brockman2016gym,towers2024gymnasium}, (2)~continuous control tasks with MuJoCo Playground~\citep{zakka2025mujoco,todorov2012mujoco}, and (3)~physics-based humanoid control in Isaac Gym~\citep{makoviychuk2021isaac}.
These tasks vary in dimensionality, reward sparsity, horizon length, and simulation environments. %

\begin{figure}[tbp]
    \centering
    \includegraphics[width=\textwidth, clip, trim=0.1cm 8cm 0.1cm 6.5cm]{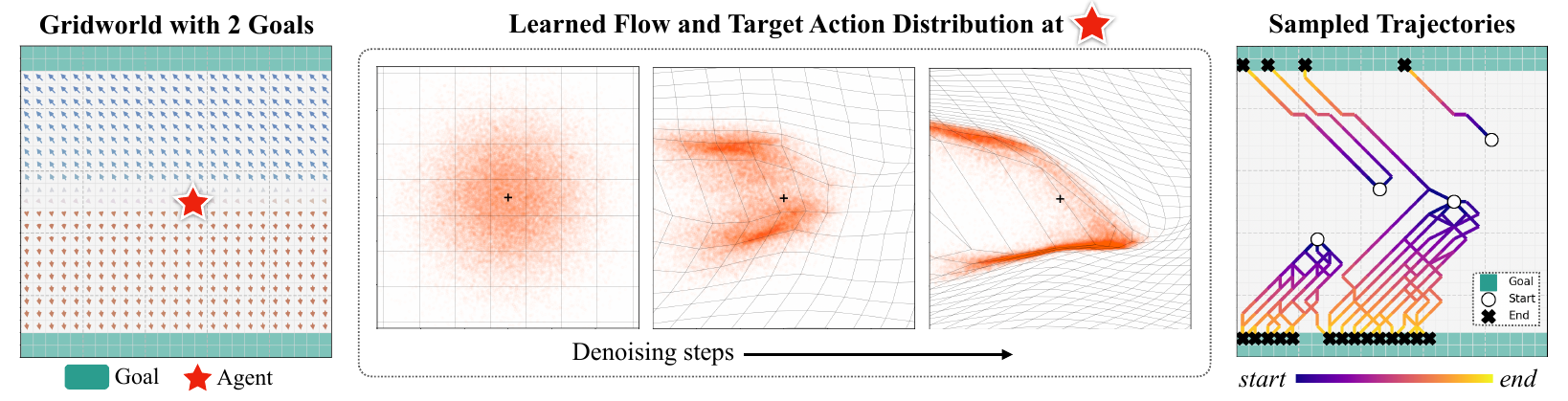}
    \caption{\textbf{Grid World}. (Left) 25$\times$25 GridWorld with green goal cells. Each arrow shows a denoised action sampled from the FPO-trained policy, conditioned on a different latent noise vector. (Center) At the saddle-point state ($\star$) shown on the left, we visualize three denoising steps $\tau$ as the initial Gaussian gradually transforms into the target distribution through the learned flow, illustrated by the deformation of the coordinate grid. (Right) Sampled trajectories from the same starting states reach different goals, illustrating the multimodal behavior captured by FPO. %
    }
    \label{fig:grid}
    \vspace{-0.0em}
\end{figure}
\subsection{GridWorld}
We first test FPO on a 25$\times$25 GridWorld environment designed to probe the policy's ability to capture multimodal action distributions. As shown in Figure~\ref{fig:grid} left, the environment consists of two high reward regions located as the top and bottom of the map (green cells). The reward is sparse: agents receive a single reward upon reaching a goal or a penalty, with no intermediate rewards. This setup creates saddle points where multiple distinct actions can lead to equally successful outcomes, offering a natural opportunity to model diverse behaviors.

We train a diffusion policy from scratch using FPO by modifying a standard implementation~\citep{yu2020ppo_for_beginners} of PPO. The policy is parameterized as a two-layer MLP modeling $p(a_t \mid s, \att)$, where $a_t \in \mathbb{R}^2$ is the action, $s\in\mathbb{R}^2$ is the grid state, and $\att \in \mathbb{R}^2$ is the latent noise vector at noise level $\tau$, initialized from $\mathcal{N}(0, I)$ at $\tau=0$. FPO consistently maximizes the return in this environment. The arrows in Figure~\ref{fig:grid} left shows denoised actions at each grid location, computed by conditioning on a random $\att \sim \mathcal{N}(0, I)$
  and running 10 steps of Euler integration. %
In Figure~\ref{fig:grid} center, we probe the learned policy by visualizing the flow over its denoising steps at the saddle point. The initial Gaussian evolves into a bimodal distribution, demonstrating that the policy captures the multi-modality of the solution at this location. Figure~\ref{fig:grid} right shows multiple trajectories sampled from the policy, initialized from various fixed starting positions. The agent exhibits multimodal behavior, with trajectories from the same starting state reaching different goals. Even when heading toward the same goal, the paths vary significantly, reflecting the policy’s ability to model diverse action sequences. 

We also train a Gaussian policy using PPO, which successfully reaches the goal regions. Compared to FPO, it exhibits more deterministic behavior, consistently favoring the nearest goal with less variation in trajectory patterns. Results are included in the supplemental material (Appendix~\ref{app:gridworld}).

\begin{figure}[!t]
    \centering
    \vspace{-1em}
    \includegraphics[width=0.99\linewidth]{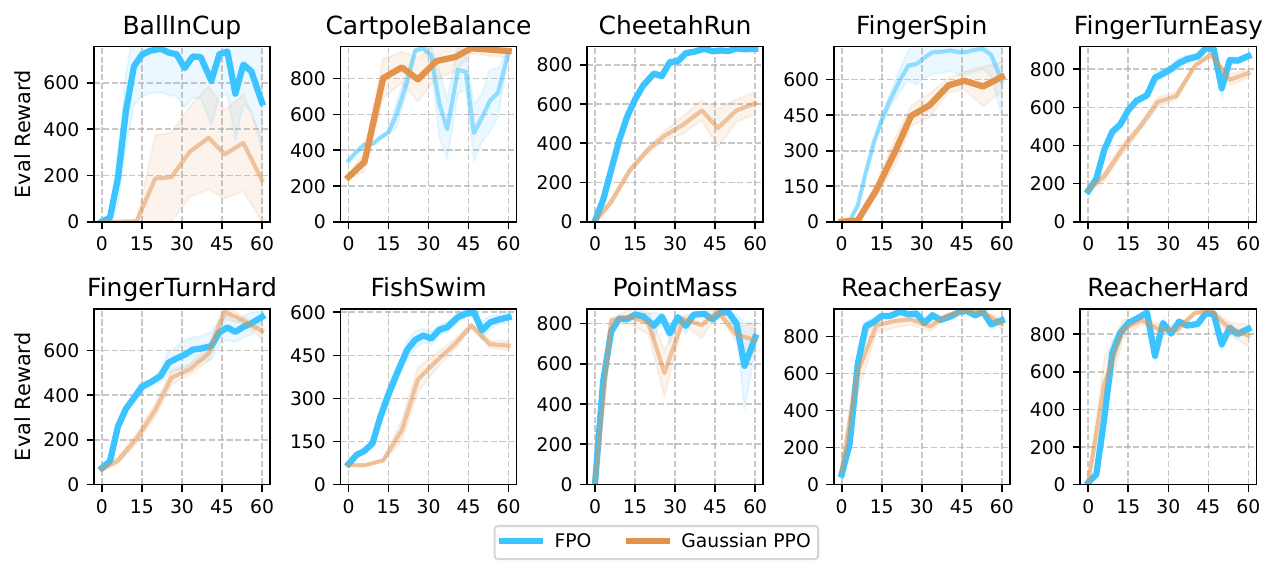}
    \vspace{-.5em}
    \caption{
        \textbf{Comparison between FPO and Gaussian PPO~\cite{schulman2017proximal} on DM Control Suite tasks.}
        Results show evaluation reward mean and standard error (y-axis) over 60M environment steps (x-axis).
        We run 5 seeds for each task; the curve with the highest terminal evaluation reward is bolded.
\textbf{}    }
    \label{fig:dm_control_gaussian}
    \vspace{.5em}
    \centering
    \includegraphics[width=0.99\linewidth]{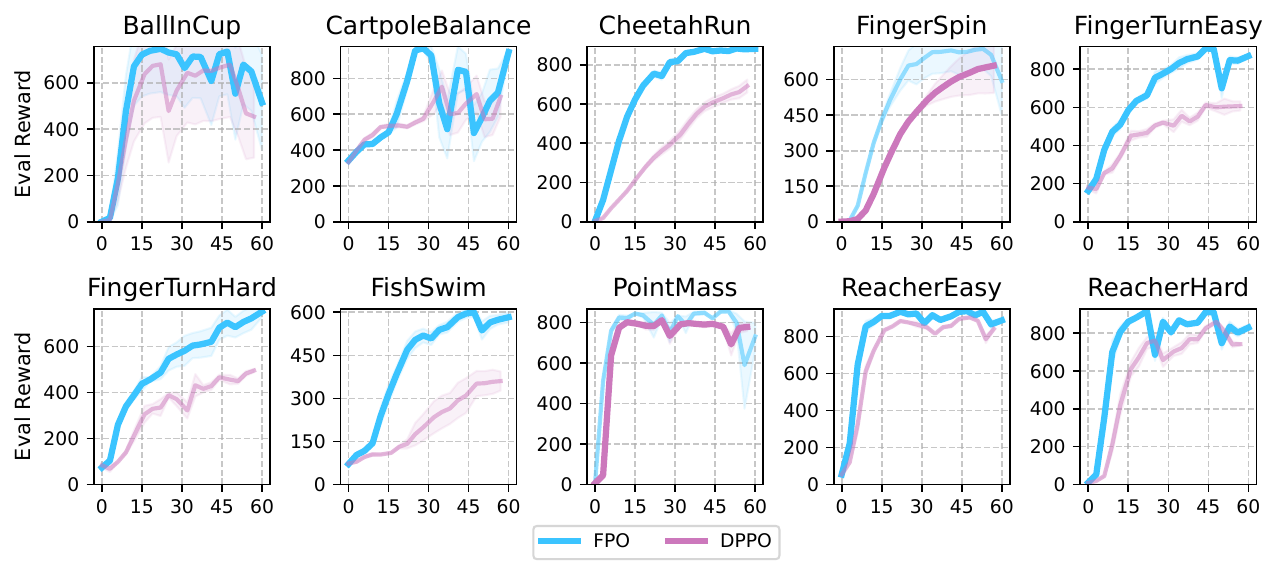}
    \vspace{-0.5em}
    \caption{
        \textbf{Comparison between FPO and DPPO~\cite{ren2024diffusion} on DM Control Suite tasks.}
        Results show evaluation reward mean and standard error (y-axis) over 60M environment steps (x-axis).
        We run 5 seeds for each task; the curve with the highest terminal evaluation reward is bolded.
    }
    \vspace{-1.2em}
    \label{fig:dm_control_dppo}
\end{figure}

\subsection{MuJoCo Playground}
\label{sec:mujoco}

Next, we evaluate FPO for continuous control using MuJoCo Playground~\cite{zakka2025mujoco}.
We compare three policy learning algorithms: (i) a Gaussian policy trained using PPO, (ii) a diffusion policy trained using FPO, and (iii) a diffusion policy trained using DPPO~\cite{ren2024diffusion}.
We evaluate these algorithms on 5 seeds for each of 10 environments adapted from the DeepMind Control Suite~\cite{tassa2018deepmind,tunyasuvunakool2020dm_control}.
Results are reported in Figures~\ref{fig:dm_control_gaussian} and ~\ref{fig:dm_control_dppo}.

\textbf{Policy implementations.}
For the Gaussian policy baseline, we run the Brax~\cite{freeman2021brax}-based implementation used by MuJoCo Playground~\cite{zakka2025mujoco}'s PPO training scripts.
We also use Brax PPO as a starting point for implementing both FPO and DPPO.
Following Section~\ref{sec:flow_policy_optimization}, only small changes are required for FPO: noisy action and timestep inputs are included as input to the policy network, Gaussian sampling is replaced with flow sampling, and the PPO loss's likelihood ratio is replaced with the FPO ratio approximation.
For DPPO, we make the same policy network modification, but apply stochastic sampling~\cite{liu2025flow} during rollouts.
We also augment each action in the experience buffer with the exact sampling path that was taken to reach it.
Following the two-layer MDP formulation in DPPO~\cite{ren2024diffusion}, we then replace intractable action likelihoods with noise-conditioned sampling path likelihoods.

\textbf{Hyperparameters.}
We match hyperparameters in Gaussian PPO, FPO, and DPPO training whenever possible: following the provided configurations in Playground~\cite{zakka2025mujoco}, all experiments use ADAM~\cite{kingma2014adam}, 60M total environment steps, batch size 1024, and 16 updates per batch.
For FPO and DPPO, we use 10 sampling steps, set learning rates to 3e-4, and swept clipping epsilon $\varepsilon^\text{clip} \in \{0.01, 0.05,0.1,0.2,0.3\}$.
For DPPO, we perturb each denoising step with Gaussian noise with standard deviation $\sigma_t$, which we sweep $\in \{0.01,0.05,0.1\}$.
We found that $\varepsilon^\text{clip}=0.05$ produces the best FPO results and $\varepsilon^\text{clip}=0.2, \sigma_t=0.05$ produced the best DPPO results; we use these values for all experiments.
For fairness, we also tuned learning rates and clipping epsilons for Gaussian PPO. We provide more details about hyperparameters and baseline tuning in Appendix~\ref{app:playground_details}.

\begin{wraptable}[19]{r}{0.35\textwidth}
    \vspace{-1.1em}
    \centering
    \begin{tabular}{ll}
        \toprule
        Method & Reward \\
        \cmidrule(r){1-1} \cmidrule(){2-2}
        Gaussian PPO & 667.8$\pm$66.0 \\
        Gaussian PPO$^\dagger$ & 577.2$\pm$74.4 \\
        DPPO & 652.5$\pm$83.7 \\
        FPO$^\ddagger$  & \textbf{759.3$\pm$45.3} \\
        \cmidrule(r){1-1} \cmidrule(l){2-2}
        FPO, 1 $(\tau,\epsilon)$ & 691.6$\pm$50.3 \\
        FPO, 4 $(\tau,\epsilon)$ & 731.2$\pm$58.2 \\
        FPO, $u$-MSE & 664.6$\pm$48.5 \\
        FPO, $\ppoeps$=0.1 & 623.3$\pm$76.3\\
        FPO, $\ppoeps$=0.2 & 526.4$\pm$76.8 \\
        \bottomrule
    \end{tabular}
    \caption{
        \textbf{FPO variant comparison.}
        We report averages and standard errors across MuJoCo tasks.
        $^\dagger$Using default hyperparameters from MuJoCo Playground.
        $^\ddagger$FPO results use 8 $(\tau, \epsilon)$ pairs, $\epsilon$-MSE, $\ppoeps=0.05$.
    }
    \label{tab:playground_ablations}
\end{wraptable}

\textbf{Results.}
We observe in Figures~\ref{fig:dm_control_gaussian} and \ref{fig:dm_control_dppo} that FPO-optimized policies outperform both Gaussian PPO and DPPO on the Playground tasks.
It outperforms both baselines in 8 of 10 tasks.

\textbf{Analysis.}
In Table~\ref{tab:playground_ablations}, we present average evaluation rewards for baselines, FPO, and several variations of FPO.
We observe:
\textbf{(1) $\boldsymbol{(\tau,\epsilon)}$ sampling is important.} 
Decreasing the number of sampled pairs generally decreases evaluation rewards.
More samples can improve learning without requiring more expensive environment steps.
\textbf{(2)} \textbf{$\epsilon$-MSE is preferable over $u$-MSE in Playground.}
$\epsilon$-MSE refers to computing flow matching losses by first converting velocity estimates to $\epsilon$ noise values;  $u$-MSE refers to MSE directly on velocity estimates.
In Playground, we found that the former produces higher average rewards.
We hypothesize that this is because $\epsilon$ scale is invariant to action scale, which results in better generalization for $\ppoeps$ choices.
For fairness, we also performed learning rate and clipping ratio sweeps for the $u$-MSE ablation.
\textbf{(3) Clipping.}
Like Gaussian PPO, the choice of $\ppoeps$ in FPO significantly impacts performance. %

\begin{table*}[t!]
\centering
\resizebox{0.9\textwidth}{!}{%
\begin{tabular}{ll|ccc}
\toprule
Methods & Goal conditioning & Success rate ($\uparrow$) & Alive duration ($\uparrow$) & MPJPE ($\downarrow$)  \\ \midrule
Gaussian PPO  & All joints        &   $\mathbf{98.7\%}$   &   $\mathbf{200.46}$  & $\mathbf{31.62}$ \\
FPO & All joints        &   $96.4\%$   &   $198.00$  & $41.98$ \\ \midrule
Gaussian PPO  & Root + Hands       &   $46.5\%$   &   $142.50$  & $97.65$  \\
FPO & Root + Hands       &   $\mathbf{70.6\%}$   &   $\mathbf{171.32}$  & $\mathbf{62.91}$\\ \midrule
Gaussian PPO  & Root              &   $29.8\%$   &   $114.06$  & $123.70$  \\
FPO & Root              &   $\mathbf{54.3\%}$   &   $\mathbf{152.90}$  & $\mathbf{73.55}$ \\ \bottomrule
\end{tabular}
}%
\caption{\textbf{Humanoid Control Quantitative Metrics.} We compare FPO with Gaussian PPO with different conditioning goals, and report the success rate, alive duration, and MPJPE averaged over all motion sequences.}
\label{tab:phc}
\end{table*}

\begin{figure}[htbp]
  \centering
  \begin{subfigure}[b]{0.33\textwidth}
    \includegraphics[width=\linewidth]{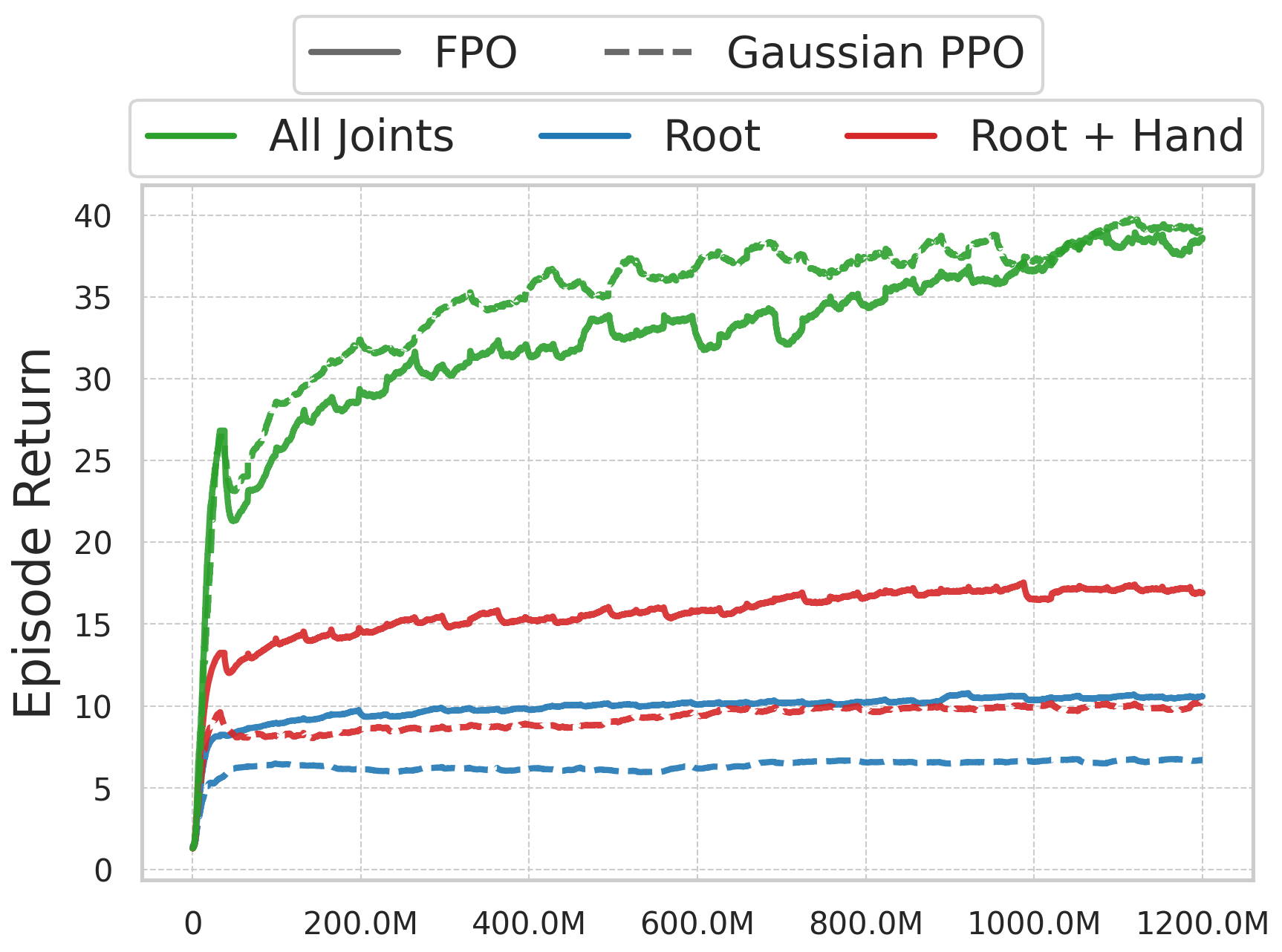}
    \caption{Episode return along training.}
    \label{fig:humanoid_sub1}
  \end{subfigure}
  \hfill
  \begin{subfigure}[b]{0.32\textwidth}
    \includegraphics[width=\linewidth]{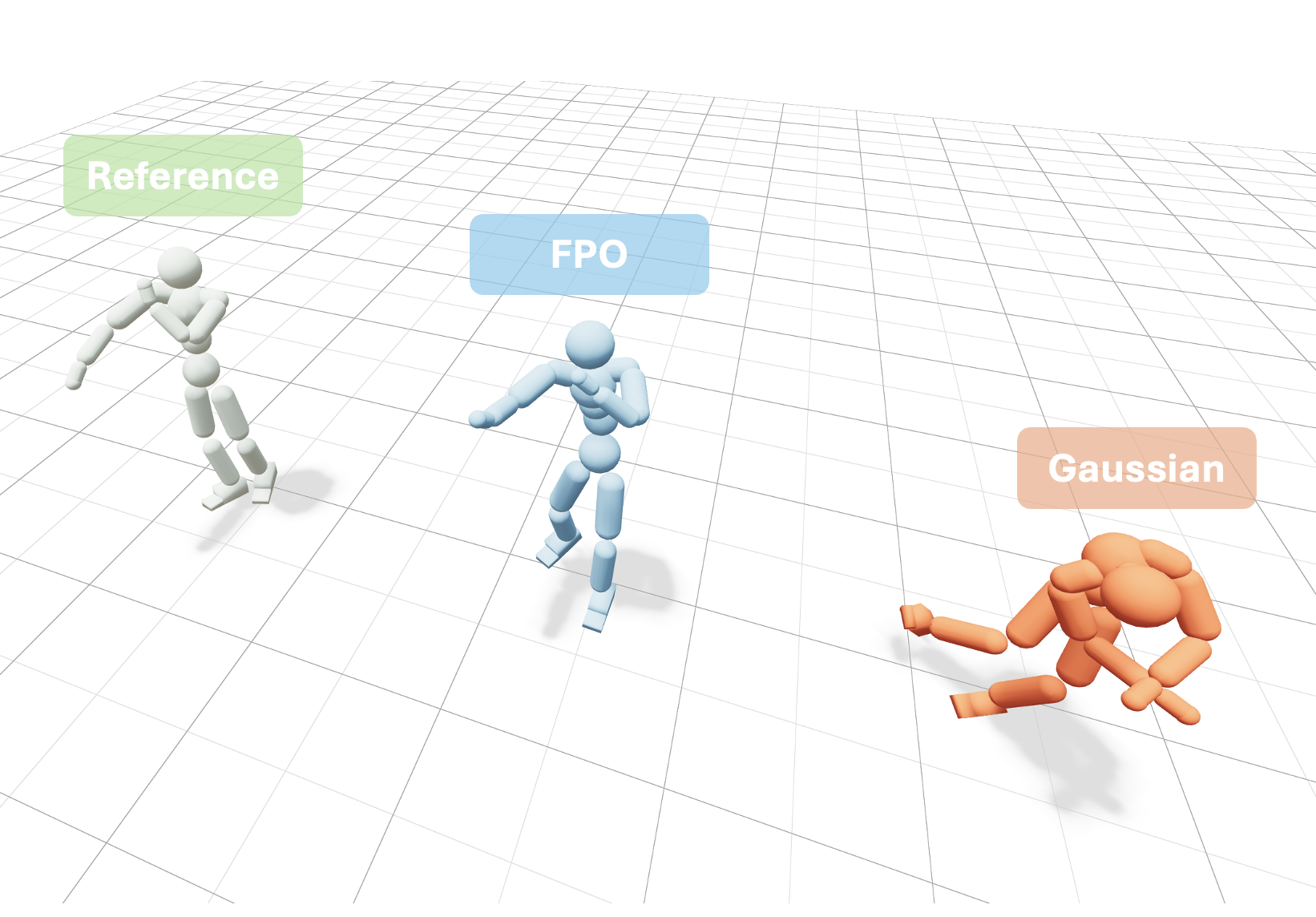}
    \caption{Root$+$hand conditioning.}
    \label{fig:humanoid_sub2}
  \end{subfigure}
  \hfill
  \begin{subfigure}[b]{0.32\textwidth}
    \includegraphics[width=\linewidth]{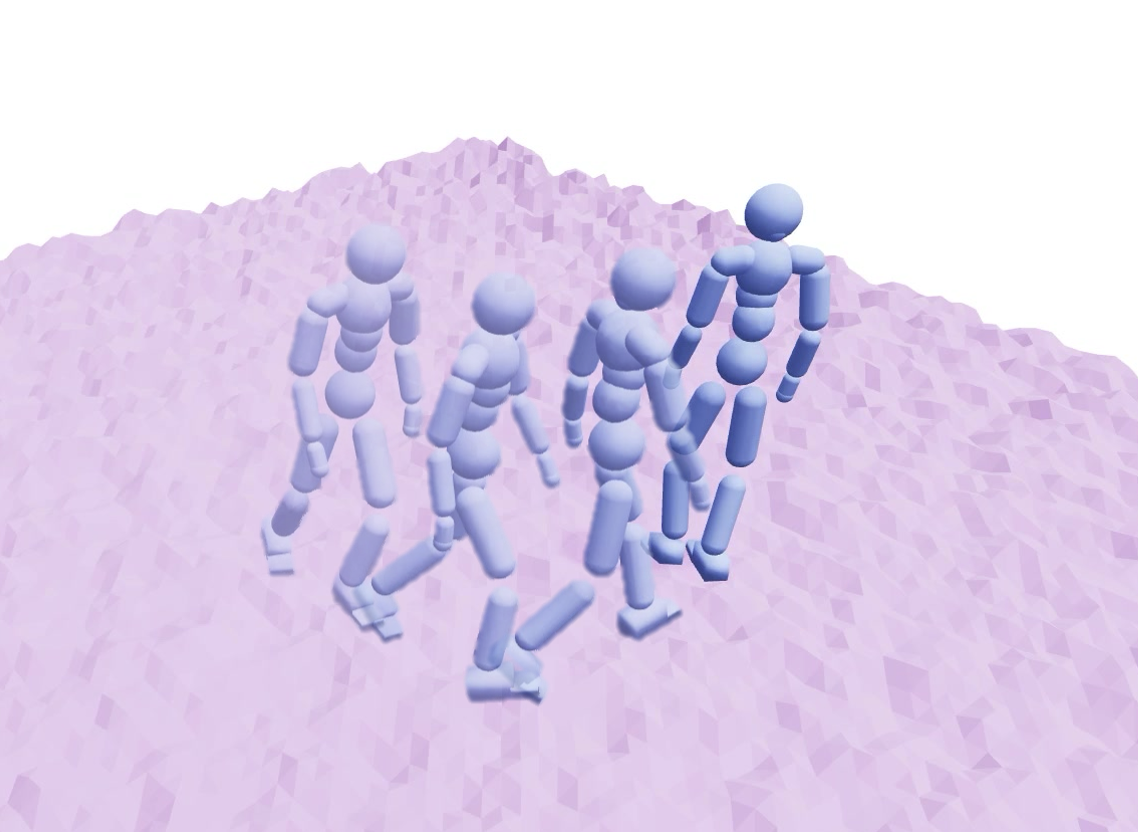}
    \caption{Rough terrain locomotion.}
    \label{fig:humanoid_sub3}
  \end{subfigure}
  \caption{\textbf{Physics-based Humanoid Control.} (a) The curves show that FPO performance is close to that of Gaussian‑PPO when conditioning on all joints and surpasses it when goals are reduced to the root or root$+$hands, indicating stronger robustness to sparse conditioning. (b) In the root$+$hands goal setting, FPO (blue) tracks the reference motion (grey) while Gaussian‑PPO (orange) falls. (c) Trained with terrain randomization, FPO walks stably across procedurally generated rough ground.}
  \label{fig:humanoid}
\end{figure}

\subsection{Humanoid Control}
Physics‑aware humanoid control is higher‑dimensional than standard MuJoCo benchmarks, making it a stringent test of FPO’s generality. We therefore train a humanoid policy to track motion‑capture (MoCap) trajectories in the PHC setting~\cite{luo2023perpetual}, using the open‑source Puffer‑PHC implementation as our baseline\footnote{\url{https://github.com/kywch/puffer-phc}}. This experiment follows the goal-conditioned imitation‑learning paradigm pioneered by DeepMimic~\cite{peng2018deepmimic}, in which simulated characters learn to reproduce reference motions. Depending on the deployment needs, these reference signals (goals) can be as rich as full‑body joint information or as sparse as root joint (pelvis) commands, providing the flexibility required for reliable sim‑to‑real transfer~\cite{videomimic}. The problem with sparse goals is under-conditioned and significantly more challenging, requiring the policy to fill in the missing joint specification in a manner that is physically plausible.

\textbf{Implementation details.} 
Our simulated agent is an SMPL‑based humanoid with 24 actuated joints, each offering six degrees of freedom and organized in a kinematic tree rooted at the pelvis, simulated in Isaac Gym~\cite{makoviychuk2021isaac}. The policy receives both proprioceptive observations and goal information computed from the motion‑capture reference.
A single policy is trained to track AMASS~\cite{mahmood2019amass} motions following PHC~\cite{luo2023perpetual}.
We use the root height, joint positions, rotations, velocity, and angular velocity in a local coordinate frame as the robot state. For goal conditioning, we compute the difference between the tracking joint information (positions, rotations, velocity, and angular velocity) and the current robot's joint information, as well as the tracking joint locations and rotations. We explore both full conditioning, \textit{i.e.,} conditioning on all joint targets, and under conditioning, \textit{i.e.,} conditioning only on the root or the root and hands targets. 
The latter matches the target signals typically provided by a joystick or VR controller.
Please note that the same imitation reward based on all joints is used for both conditioning experiments.
The per‑joint tracking reward is computed as in DeepMimic~\cite{peng2018deepmimic}.

\textbf{Evaluation.} For evaluation, we compute the success rate, considering an imitation unsuccessful if the average distance between the body joints and the reference motion exceeds 0.5 meters at any point during the sequence. We also report the average duration the agent stays alive till it completes the tracking or falls. Finally, we compute the global mean per-joint position error (MPJPE) on the conditioned goals.

\textbf{Results.} 
Figure~\ref{fig:humanoid_sub1} shows that we successfully train FPO from scratch on this high-dimensional control task. With full joint conditioning, FPO performance is close to Gaussian PPO. However, when the model is under-conditioned---e.g., conditioned only on the root or the root and hands---FPO outperforms Gaussian PPO, highlighting the advantage of flow-based policies. 
While prior methods can also achieve sparse-goal control, they often rely on training a teacher policy that conditions on full joint reference first and then distilling the knowledge to sparse conditioned policies~\cite{tessler2024maskedmimic,videomimic,li2025clone} or training a separate encoder observing sparse references~\cite{luo2023universal,luo2024omnigrasp}.

Figure~\ref{fig:humanoid_sub2} visualizes the behaviors in the root$+$hands setting (left-to‑right: reference motion, FPO, Gaussian‑PPO); FPO tracks the target closely, whereas the Gaussian policy drifts. Table~\ref{tab:phc} quantifies these trends, with FPO achieving much higher success rates in the under‑conditioned scenarios.
Finally, as illustrated in Fig.~\ref{fig:humanoid_sub3}, FPO trained with terrain randomization enables the humanoid to traverse rough terrain, showing potential for sim‑to‑real transfer. Please see the supplemental video for more qualitative results.

\section{Discussion and Limitations} \label{sec:limitations}
We introduce Flow Policy Optimization (FPO), an algorithm for training flow-based generative models using policy gradients. FPO reformulates policy optimization as minimizing an advantage-weighted conditional flow matching (CFM) objective,  enabling stable training without requiring explicit likelihood computation. %
It integrates easily with PPO-style algorithms, and crucially, preserves the flow-based structure of the policy—allowing the resulting model to be used with standard flow-based mechanisms such as sampling, distillation, and fine-tuning. We demonstrate FPO across a range of control tasks, including a challenging humanoid setting where it enables training from scratch under sparse goal conditioning, where Gaussian policies fail to learn.

The training and deployment of flow-based policies is generally more computationally intensive than for corresponding Gaussian policies.
FPO also lacks established machinery such as KL divergence estimation for adaptive learning rates and entropy regularization. 

We also explored applying FPO to fine-tune a pre-trained image diffusion model using reinforcement learning. While promising in principle, we found this setting to be unstable in practice---likely  due to the issue of fine-tuning diffusion models on its own output multiple times as noted in recent works~\cite{shumailov2024ai,shumailov2023curse,alemohammad2024self}. In particular, we observed sensitivity to classifier-free guidance (CFG) that compounds with self-generated data, even outside of the RL framework. This suggests that the instability is not a limitation of FPO itself, but a broader challenge in applying reinforcement learning to image generation. Please see the supplementary material for more detail. %

Despite these limitations, FPO offers a simple and flexible bridge between flow-based models and online reinforcement learning. We are  particularly excited to see future work apply FPO in settings where flow-based policies are already pretrained---such as behavior-cloned diffusion policies in robotics---where FPO’s compatibility and simplicity may offer practical benefits for fine-tuning with task reward.

\subsection*{Acknowledgments}

We thank Qiyang (Colin) Li, Oleg Rybkin, Lily Goli and Michael Psenka for helpful discussions and feedback on the manuscript. We thank Arthur Allshire, Tero Karras, Miika Aittala, Kevin Zakka and Seohong Park for insightful input and feedback on implementation details and the broader context of this work. This project was funded in part by NSF:CNS-2235013, IARPA DOI/IBC No. 140D0423C0035, and Bakar fellows. CK and BY are supported by NSF fellowship. SG is supported by the NVIDIA Graduate Fellowship

\clearpage

\bibliographystyle{unsrtnat} %
\bibliography{references}

\begin{thebibliography}{84}
\providecommand{\natexlab}[1]{#1}
\providecommand{\url}[1]{\texttt{#1}}
\expandafter\ifx\csname urlstyle\endcsname\relax
  \providecommand{\doi}[1]{doi: #1}\else
  \providecommand{\doi}{doi: \begingroup \urlstyle{rm}\Url}\fi

\bibitem[Ramesh et~al.(2022)Ramesh, Dhariwal, Nichol, Chu, and Chen]{ramesh2022hierarchical}
Aditya Ramesh, Prafulla Dhariwal, Alex Nichol, Casey Chu, and Mark Chen.
\newblock Hierarchical text-conditional image generation with clip latents.
\newblock \emph{arXiv preprint arXiv:2204.06125}, 2022.

\bibitem[Saharia et~al.(2022)Saharia, Chan, Saxena, Li, Whang, Denton, Ghasemipour, Gontijo~Lopes, Karagol~Ayan, Salimans, et~al.]{saharia2022photorealistic}
Chitwan Saharia, William Chan, Saurabh Saxena, Lala Li, Jay Whang, Emily~L Denton, Kamyar Ghasemipour, Raphael Gontijo~Lopes, Burcu Karagol~Ayan, Tim Salimans, et~al.
\newblock Photorealistic text-to-image diffusion models with deep language understanding.
\newblock 2022.

\bibitem[Ho et~al.(2022{\natexlab{a}})Ho, Chan, Saharia, Whang, Gao, Gritsenko, Kingma, Poole, Norouzi, Fleet, et~al.]{ho2022imagen}
Jonathan Ho, William Chan, Chitwan Saharia, Jay Whang, Ruiqi Gao, Alexey Gritsenko, Diederik~P Kingma, Ben Poole, Mohammad Norouzi, David~J Fleet, et~al.
\newblock Imagen video: High definition video generation with diffusion models.
\newblock \emph{arXiv preprint arXiv:2210.02303}, 2022{\natexlab{a}}.

\bibitem[Brooks et~al.(2024)Brooks, Peebles, Holmes, DePue, Guo, Jing, Schnurr, Taylor, Luhman, Luhman, Ng, Wang, and Ramesh]{videoworldsimulators2024}
Tim Brooks, Bill Peebles, Connor Holmes, Will DePue, Yufei Guo, Li~Jing, David Schnurr, Joe Taylor, Troy Luhman, Eric Luhman, Clarence Ng, Ricky Wang, and Aditya Ramesh.
\newblock Video generation models as world simulators.
\newblock 2024.
\newblock URL \url{https://openai.com/research/video-generation-models-as-world-simulators}.

\bibitem[Polyak et~al.(2024)Polyak, Zohar, Brown, Tjandra, Sinha, Lee, Vyas, Shi, Ma, Chuang, et~al.]{polyak2024movie}
Adam Polyak, Amit Zohar, Andrew Brown, Andros Tjandra, Animesh Sinha, Ann Lee, Apoorv Vyas, Bowen Shi, Chih-Yao Ma, Ching-Yao Chuang, et~al.
\newblock Movie gen: A cast of media foundation models.
\newblock \emph{arXiv preprint arXiv:2410.13720}, 2024.

\bibitem[Veo-Team et~al.(2024)Veo-Team, :, Gupta, Razavi, Toor, Gupta, Erhan, Shaw, Lau, Belletti, Barth-Maron, Shaw, Erdogan, Sidahmed, Nandwani, Moraldo, Kim, Blok, Donahue, Lezama, Mathewson, David, Lorrain, van Zee, Narasimhan, Wang, Babaeizadeh, Papalampidi, Pezzotti, Jha, Barnes, Kindermans, Hornung, Villegas, Poplin, Zaiem, Dieleman, Ebrahimi, Wisdom, Zhang, Fruchter, Nørly, Hua, Yan, Du, and Chen]{veo2}
Veo-Team, :, Agrim Gupta, Ali Razavi, Andeep Toor, Ankush Gupta, Dumitru Erhan, Eleni Shaw, Eric Lau, Frank Belletti, Gabe Barth-Maron, Gregory Shaw, Hakan Erdogan, Hakim Sidahmed, Henna Nandwani, Hernan Moraldo, Hyunjik Kim, Irina Blok, Jeff Donahue, José Lezama, Kory Mathewson, Kurtis David, Matthieu~Kim Lorrain, Marc van Zee, Medhini Narasimhan, Miaosen Wang, Mohammad Babaeizadeh, Nelly Papalampidi, Nick Pezzotti, Nilpa Jha, Parker Barnes, Pieter-Jan Kindermans, Rachel Hornung, Ruben Villegas, Ryan Poplin, Salah Zaiem, Sander Dieleman, Sayna Ebrahimi, Scott Wisdom, Serena Zhang, Shlomi Fruchter, Signe Nørly, Weizhe Hua, Xinchen Yan, Yuqing Du, and Yutian Chen.
\newblock Veo 2.
\newblock 2024.
\newblock URL \url{https://deepmind.google/technologies/veo/veo-2/}.

\bibitem[Liu et~al.(2023)Liu, Chen, Yuan, Mei, Liu, Mandic, Wang, and Plumbley]{liu2023audioldmtexttoaudiogenerationlatent}
Haohe Liu, Zehua Chen, Yi~Yuan, Xinhao Mei, Xubo Liu, Danilo Mandic, Wenwu Wang, and Mark~D. Plumbley.
\newblock Audioldm: Text-to-audio generation with latent diffusion models, 2023.
\newblock URL \url{https://arxiv.org/abs/2301.12503}.

\bibitem[Kong et~al.(2021)Kong, Ping, Huang, Zhao, and Catanzaro]{kong2021diffwaveversatilediffusionmodel}
Zhifeng Kong, Wei Ping, Jiaji Huang, Kexin Zhao, and Bryan Catanzaro.
\newblock Diffwave: A versatile diffusion model for audio synthesis, 2021.
\newblock URL \url{https://arxiv.org/abs/2009.09761}.

\bibitem[Chi et~al.(2024{\natexlab{a}})Chi, Xu, Feng, Cousineau, Du, Burchfiel, Tedrake, and Song]{chi2024diffusionpolicy}
Cheng Chi, Zhenjia Xu, Siyuan Feng, Eric Cousineau, Yilun Du, Benjamin Burchfiel, Russ Tedrake, and Shuran Song.
\newblock Diffusion policy: Visuomotor policy learning via action diffusion.
\newblock \emph{The International Journal of Robotics Research}, 2024{\natexlab{a}}.

\bibitem[Raja et~al.(2025)Raja, {\v{S}}{\'\i}pka, Psenka, Kreiman, Pavelka, and Krishnapriyan]{raja2025action}
Sanjeev Raja, Martin {\v{S}}{\'\i}pka, Michael Psenka, Tobias Kreiman, Michal Pavelka, and Aditi~S Krishnapriyan.
\newblock Action-minimization meets generative modeling: Efficient transition path sampling with the onsager-machlup functional.
\newblock \emph{arXiv preprint arXiv:2504.18506}, 2025.

\bibitem[Chu et~al.(2025)Chu, Zhai, Yang, Tong, Xie, Schuurmans, Le, Levine, and Ma]{chu2025sft}
Tianzhe Chu, Yuexiang Zhai, Jihan Yang, Shengbang Tong, Saining Xie, Dale Schuurmans, Quoc~V Le, Sergey Levine, and Yi~Ma.
\newblock Sft memorizes, rl generalizes: A comparative study of foundation model post-training.
\newblock \emph{arXiv preprint arXiv:2501.17161}, 2025.

\bibitem[Liu et~al.(2024)Liu, Feng, Wang, Wang, Liu, Zhao, Dengr, Ruan, Dai, Guo, et~al.]{liu2024deepseek}
Aixin Liu, Bei Feng, Bin Wang, Bingxuan Wang, Bo~Liu, Chenggang Zhao, Chengqi Dengr, Chong Ruan, Damai Dai, Daya Guo, et~al.
\newblock Deepseek-v2: A strong, economical, and efficient mixture-of-experts language model.
\newblock \emph{arXiv preprint arXiv:2405.04434}, 2024.

\bibitem[Lipman et~al.(2023)Lipman, Chen, Ben-Hamu, Nickel, and Le]{lipman2023flowmatchinggenerativemodeling}
Yaron Lipman, Ricky T.~Q. Chen, Heli Ben-Hamu, Maximilian Nickel, and Matt Le.
\newblock Flow matching for generative modeling, 2023.
\newblock URL \url{https://arxiv.org/abs/2210.02747}.

\bibitem[Zakka et~al.(2025)Zakka, Tabanpour, Liao, Haiderbhai, Holt, Luo, Allshire, Frey, Sreenath, Kahrs, et~al.]{zakka2025mujoco}
Kevin Zakka, Baruch Tabanpour, Qiayuan Liao, Mustafa Haiderbhai, Samuel Holt, Jing~Yuan Luo, Arthur Allshire, Erik Frey, Koushil Sreenath, Lueder~A Kahrs, et~al.
\newblock Mujoco playground.
\newblock \emph{arXiv preprint arXiv:2502.08844}, 2025.

\bibitem[Sutton et~al.(1999)Sutton, McAllester, Singh, and Mansour]{sutton1999policygradient}
Richard~S. Sutton, David McAllester, Satinder~P. Singh, and Yishay Mansour.
\newblock Policy gradient methods for reinforcement learning with function approximation.
\newblock In \emph{Proceedings of the 12th International Conference on Neural Information Processing Systems (NeurIPS)}, pages 1057--1063, 1999.

\bibitem[Williams(1992)]{williams1992simple}
Ronald~J Williams.
\newblock Simple statistical gradient-following algorithms for connectionist reinforcement learning.
\newblock \emph{Machine learning}, 1992.

\bibitem[Kakade(2002)]{kakade2002naturalpg}
Sham~M. Kakade.
\newblock A natural policy gradient.
\newblock In \emph{Proceedings of the 14th International Conference on Neural Information Processing Systems (NeurIPS)}, pages 1531--1538, 2002.

\bibitem[Peters and Schaal(2008)]{peters2008nac}
Jan Peters and Stefan Schaal.
\newblock Natural actor–critic.
\newblock \emph{Neurocomputing}, 71\penalty0 (7–9):\penalty0 1180--1190, 2008.

\bibitem[Schulman et~al.(2015{\natexlab{a}})Schulman, Levine, Abbeel, Jordan, and Moritz]{schulman2015trust}
John Schulman, Sergey Levine, Pieter Abbeel, Michael Jordan, and Philipp Moritz.
\newblock Trust region policy optimization.
\newblock In \emph{International conference on machine learning}, pages 1889--1897. PMLR, 2015{\natexlab{a}}.

\bibitem[Schulman et~al.(2017)Schulman, Wolski, Dhariwal, Radford, and Klimov]{schulman2017proximal}
John Schulman, Filip Wolski, Prafulla Dhariwal, Alec Radford, and Oleg Klimov.
\newblock Proximal policy optimization algorithms.
\newblock \emph{arXiv preprint arXiv:1707.06347}, 2017.

\bibitem[Mnih et~al.(2016)Mnih, Badia, Mirza, Graves, Harley, Lillicrap, Silver, and Kavukcuoglu]{mnih2016a3c}
Volodymyr Mnih, Adria~Puigdomenech Badia, Mehdi Mirza, Alex Graves, Tim Harley, Timothy Lillicrap, David Silver, and Koray Kavukcuoglu.
\newblock Asynchronous methods for deep reinforcement learning.
\newblock In \emph{Proceedings of the 33rd International Conference on Machine Learning (ICML)}, pages 1928--1937, 2016.

\bibitem[Wang et~al.(2016)Wang, Schaul, Hessel, Hasselt, Lanctot, and de~Freitas]{wang2016acer}
Ziyu Wang, Tom Schaul, Matteo Hessel, Hado Hasselt, Marc Lanctot, and Nando de~Freitas.
\newblock Sample efficient actor–critic with experience replay.
\newblock In \emph{Proceedings of the 30th International Conference on Neural Information Processing Systems (NeurIPS)}, pages 1061--1071, 2016.

\bibitem[Shao et~al.(2024)Shao, Wang, Zhu, Xu, Song, Bi, Zhang, Zhang, Li, Wu, et~al.]{shao2024deepseekmath}
Zhihong Shao, Peiyi Wang, Qihao Zhu, Runxin Xu, Junxiao Song, Xiao Bi, Haowei Zhang, Mingchuan Zhang, YK~Li, Y~Wu, et~al.
\newblock Deepseekmath: Pushing the limits of mathematical reasoning in open language models.
\newblock \emph{arXiv preprint arXiv:2402.03300}, 2024.

\bibitem[Duan et~al.(2016)Duan, Chen, Houthooft, Schulman, and Abbeel]{duan2016benchmarking}
Yan Duan, Xi~Chen, Rein Houthooft, John Schulman, and Pieter Abbeel.
\newblock Benchmarking deep reinforcement learning for continuous control.
\newblock In \emph{International conference on machine learning}, pages 1329--1338. PMLR, 2016.

\bibitem[Huang et~al.(2024)Huang, Gallou{\'e}dec, Felten, Raffin, Dossa, Zhao, Sullivan, Makoviychuk, Makoviichuk, Danesh, et~al.]{huang2024open}
Shengyi Huang, Quentin Gallou{\'e}dec, Florian Felten, Antonin Raffin, Rousslan Fernand~Julien Dossa, Yanxiao Zhao, Ryan Sullivan, Viktor Makoviychuk, Denys Makoviichuk, Mohamad~H Danesh, et~al.
\newblock Open rl benchmark: Comprehensive tracked experiments for reinforcement learning.
\newblock \emph{arXiv preprint arXiv:2402.03046}, 2024.

\bibitem[Rudin et~al.(2022)Rudin, Hoeller, Reist, and Hutter]{rudin2022learning}
Nikita Rudin, David Hoeller, Philipp Reist, and Marco Hutter.
\newblock Learning to walk in minutes using massively parallel deep reinforcement learning.
\newblock In \emph{Proceedings of the 5th Conference on Robot Learning}, volume 164 of \emph{Proceedings of Machine Learning Research}, pages 91--100. PMLR, 2022.
\newblock URL \url{https://proceedings.mlr.press/v164/rudin22a.html}.

\bibitem[Schwarke et~al.(2023)Schwarke, Klemm, Boon, Bjelonic, and Hutter]{schwarke2023curiosity}
Clemens Schwarke, Victor Klemm, Matthijs van~der Boon, Marko Bjelonic, and Marco Hutter.
\newblock Curiosity-driven learning of joint locomotion and manipulation tasks.
\newblock In \emph{Proceedings of The 7th Conference on Robot Learning}, volume 229 of \emph{Proceedings of Machine Learning Research}, pages 2594--2610. PMLR, 2023.
\newblock URL \url{https://proceedings.mlr.press/v229/schwarke23a.html}.

\bibitem[Mittal et~al.(2024)Mittal, Rudin, Klemm, Allshire, and Hutter]{mittal2024symmetry}
Mayank Mittal, Nikita Rudin, Victor Klemm, Arthur Allshire, and Marco Hutter.
\newblock Symmetry considerations for learning task symmetric robot policies.
\newblock In \emph{2024 IEEE International Conference on Robotics and Automation (ICRA)}, pages 7433--7439, 2024.
\newblock \doi{10.1109/ICRA57147.2024.10611493}.

\bibitem[Allshire et~al.(2025)Allshire, Choi, Zhang, McAllister, Zhang, Kim, Darrell, Abbeel, Malik, and Kanazawa]{videomimic}
Arthur Allshire, Hongsuk Choi, Junyi Zhang, David McAllister, Anthony Zhang, Chung~Min Kim, Trevor Darrell, Pieter Abbeel, Jitendra Malik, and Angjoo Kanazawa.
\newblock Visual imitation enables contextual humanoid control.
\newblock \emph{arXiv preprint arXiv:2505.03729}, 2025.

\bibitem[Akkaya et~al.(2019)Akkaya, Andrychowicz, Chociej, Litwin, McGrew, Petron, Paino, Plappert, Powell, Ribas, et~al.]{akkaya2019solving}
Ilge Akkaya, Marcin Andrychowicz, Maciek Chociej, Mateusz Litwin, Bob McGrew, Arthur Petron, Alex Paino, Matthias Plappert, Glenn Powell, Raphael Ribas, et~al.
\newblock Solving rubik's cube with a robot hand.
\newblock \emph{arXiv preprint arXiv:1910.07113}, 2019.

\bibitem[Chen et~al.(2021{\natexlab{a}})Chen, Xu, and Agrawal]{chen2021system}
Tao Chen, Jie Xu, and Pulkit Agrawal.
\newblock A system for general in-hand object re-orientation.
\newblock \emph{Conference on Robot Learning}, 2021{\natexlab{a}}.

\bibitem[Qi et~al.(2023)Qi, Yi, Suresh, Lambeta, Ma, Calandra, and Malik]{qi2023general}
Haozhi Qi, Brent Yi, Sudharshan Suresh, Mike Lambeta, Yi~Ma, Roberto Calandra, and Jitendra Malik.
\newblock General in-hand object rotation with vision and touch.
\newblock In \emph{Conference on Robot Learning}, pages 2549--2564. PMLR, 2023.

\bibitem[Qi et~al.(2025)Qi, Yi, Lambeta, Ma, Calandra, and Malik]{qi2025simple}
Haozhi Qi, Brent Yi, Mike Lambeta, Yi~Ma, Roberto Calandra, and Jitendra Malik.
\newblock From simple to complex skills: The case of in-hand object reorientation.
\newblock \emph{arXiv preprint arXiv:2501.05439}, 2025.

\bibitem[Ouyang et~al.(2022)Ouyang, Wu, Jiang, Almeida, Wainwright, Mishkin, Zhang, Agarwal, Slama, Ray, et~al.]{ouyang2022training}
Long Ouyang, Jeffrey Wu, Xu~Jiang, Diogo Almeida, Carroll Wainwright, Pamela Mishkin, Chong Zhang, Sandhini Agarwal, Katarina Slama, Alex Ray, et~al.
\newblock Training language models to follow instructions with human feedback.
\newblock \emph{Advances in neural information processing systems}, 2022.

\bibitem[Christiano et~al.(2023)Christiano, Leike, Brown, Martic, Legg, and Amodei]{christiano2023deepreinforcementlearninghuman}
Paul Christiano, Jan Leike, Tom~B. Brown, Miljan Martic, Shane Legg, and Dario Amodei.
\newblock Deep reinforcement learning from human preferences, 2023.
\newblock URL \url{https://arxiv.org/abs/1706.03741}.

\bibitem[DeepSeek-AI et~al.(2025)DeepSeek-AI, Guo, Yang, Zhang, Song, Zhang, Xu, Zhu, Ma, Wang, Bi, Zhang, Yu, Wu, Wu, Gou, Shao, Li, Gao, Liu, Xue, Wang, Wu, Feng, Lu, Zhao, Deng, Zhang, Ruan, Dai, Chen, Ji, Li, Lin, Dai, Luo, Hao, Chen, Li, Zhang, Bao, Xu, Wang, Ding, Xin, Gao, Qu, Li, Guo, Li, Wang, Chen, Yuan, Qiu, Li, Cai, Ni, Liang, Chen, Dong, Hu, Gao, Guan, Huang, Yu, Wang, Zhang, Zhao, Wang, Zhang, Xu, Xia, Zhang, Zhang, Tang, Li, Wang, Li, Tian, Huang, Zhang, Wang, Chen, Du, Ge, Zhang, Pan, Wang, Chen, Jin, Chen, Lu, Zhou, Chen, Ye, Wang, Yu, Zhou, Pan, Li, Zhou, Wu, Ye, Yun, Pei, Sun, Wang, Zeng, Zhao, Liu, Liang, Gao, Yu, Zhang, Xiao, An, Liu, Wang, Chen, Nie, Cheng, Liu, Xie, Liu, Yang, Li, Su, Lin, Li, Jin, Shen, Chen, Sun, Wang, Song, Zhou, Wang, Shan, Li, Wang, Wei, Zhang, Xu, Li, Zhao, Sun, Wang, Yu, Zhang, Shi, Xiong, He, Piao, Wang, Tan, Ma, Liu, Guo, Ou, Wang, Gong, Zou, He, Xiong, Luo, You, Liu, Zhou, Zhu, Xu, Huang, Li, Zheng, Zhu, Ma, Tang, Zha, Yan, Ren, Ren, Sha, Fu, Xu, Xie, Zhang,
  Hao, Ma, Yan, Wu, Gu, Zhu, Liu, Li, Xie, Song, Pan, Huang, Xu, Zhang, and Zhang]{deepseekai2025deepseekr1incentivizingreasoningcapability}
DeepSeek-AI, Daya Guo, Dejian Yang, Haowei Zhang, Junxiao Song, Ruoyu Zhang, Runxin Xu, Qihao Zhu, Shirong Ma, Peiyi Wang, Xiao Bi, Xiaokang Zhang, Xingkai Yu, Yu~Wu, Z.~F. Wu, Zhibin Gou, Zhihong Shao, Zhuoshu Li, Ziyi Gao, Aixin Liu, Bing Xue, Bingxuan Wang, Bochao Wu, Bei Feng, Chengda Lu, Chenggang Zhao, Chengqi Deng, Chenyu Zhang, Chong Ruan, Damai Dai, Deli Chen, Dongjie Ji, Erhang Li, Fangyun Lin, Fucong Dai, Fuli Luo, Guangbo Hao, Guanting Chen, Guowei Li, H.~Zhang, Han Bao, Hanwei Xu, Haocheng Wang, Honghui Ding, Huajian Xin, Huazuo Gao, Hui Qu, Hui Li, Jianzhong Guo, Jiashi Li, Jiawei Wang, Jingchang Chen, Jingyang Yuan, Junjie Qiu, Junlong Li, J.~L. Cai, Jiaqi Ni, Jian Liang, Jin Chen, Kai Dong, Kai Hu, Kaige Gao, Kang Guan, Kexin Huang, Kuai Yu, Lean Wang, Lecong Zhang, Liang Zhao, Litong Wang, Liyue Zhang, Lei Xu, Leyi Xia, Mingchuan Zhang, Minghua Zhang, Minghui Tang, Meng Li, Miaojun Wang, Mingming Li, Ning Tian, Panpan Huang, Peng Zhang, Qiancheng Wang, Qinyu Chen, Qiushi Du, Ruiqi Ge, Ruisong
  Zhang, Ruizhe Pan, Runji Wang, R.~J. Chen, R.~L. Jin, Ruyi Chen, Shanghao Lu, Shangyan Zhou, Shanhuang Chen, Shengfeng Ye, Shiyu Wang, Shuiping Yu, Shunfeng Zhou, Shuting Pan, S.~S. Li, Shuang Zhou, Shaoqing Wu, Shengfeng Ye, Tao Yun, Tian Pei, Tianyu Sun, T.~Wang, Wangding Zeng, Wanjia Zhao, Wen Liu, Wenfeng Liang, Wenjun Gao, Wenqin Yu, Wentao Zhang, W.~L. Xiao, Wei An, Xiaodong Liu, Xiaohan Wang, Xiaokang Chen, Xiaotao Nie, Xin Cheng, Xin Liu, Xin Xie, Xingchao Liu, Xinyu Yang, Xinyuan Li, Xuecheng Su, Xuheng Lin, X.~Q. Li, Xiangyue Jin, Xiaojin Shen, Xiaosha Chen, Xiaowen Sun, Xiaoxiang Wang, Xinnan Song, Xinyi Zhou, Xianzu Wang, Xinxia Shan, Y.~K. Li, Y.~Q. Wang, Y.~X. Wei, Yang Zhang, Yanhong Xu, Yao Li, Yao Zhao, Yaofeng Sun, Yaohui Wang, Yi~Yu, Yichao Zhang, Yifan Shi, Yiliang Xiong, Ying He, Yishi Piao, Yisong Wang, Yixuan Tan, Yiyang Ma, Yiyuan Liu, Yongqiang Guo, Yuan Ou, Yuduan Wang, Yue Gong, Yuheng Zou, Yujia He, Yunfan Xiong, Yuxiang Luo, Yuxiang You, Yuxuan Liu, Yuyang Zhou, Y.~X. Zhu,
  Yanhong Xu, Yanping Huang, Yaohui Li, Yi~Zheng, Yuchen Zhu, Yunxian Ma, Ying Tang, Yukun Zha, Yuting Yan, Z.~Z. Ren, Zehui Ren, Zhangli Sha, Zhe Fu, Zhean Xu, Zhenda Xie, Zhengyan Zhang, Zhewen Hao, Zhicheng Ma, Zhigang Yan, Zhiyu Wu, Zihui Gu, Zijia Zhu, Zijun Liu, Zilin Li, Ziwei Xie, Ziyang Song, Zizheng Pan, Zhen Huang, Zhipeng Xu, Zhongyu Zhang, and Zhen Zhang.
\newblock Deepseek-r1: Incentivizing reasoning capability in llms via reinforcement learning, 2025.
\newblock URL \url{https://arxiv.org/abs/2501.12948}.

\bibitem[Mistral-AI et~al.(2025)Mistral-AI, :, Rastogi, Jiang, Lo, Berrada, Lample, Rute, Barmentlo, Yadav, Khandelwal, Chandu, Blier, Saulnier, Dinot, Darrin, Gupta, Soletskyi, Vaze, Scao, Wang, Yang, Liu, Sablayrolles, Héliou, Martin, Ehrenberg, Agarwal, Roux, Darcet, Mensch, Bout, Rozière, Monicault, Bamford, Wallenwein, Renaudin, Lanfranchi, Dabert, Mizelle, de~las Casas, Chane-Sane, Fugier, Hanna, Delerce, Guinet, Novikov, Martin, Jaju, Ludziejewski, Chabran, Delignon, Studnia, Amar, Roberts, Denize, Saxena, Jain, Zhao, Martin, Gao, Lavaud, Pellat, Guillaumin, Felardos, Augustin, Seznec, Raghuraman, Duchenne, Wang, von Platen, Saffer, Jacob, Wambergue, Kurylowicz, Muddireddy, Chagniot, Stock, Agrawal, Sauvestre, Delacourt, Gandhi, Subramanian, Dalal, Gandhi, Ghosh, Mishra, Aithal, Antoniak, Schueller, Lavril, Robert, Wang, Lacroix, Nemychnikova, Paltz, Richard, Li, Marshall, Zhang, and Tang]{mistralai2025magistral}
Mistral-AI, :, Abhinav Rastogi, Albert~Q. Jiang, Andy Lo, Gabrielle Berrada, Guillaume Lample, Jason Rute, Joep Barmentlo, Karmesh Yadav, Kartik Khandelwal, Khyathi~Raghavi Chandu, Léonard Blier, Lucile Saulnier, Matthieu Dinot, Maxime Darrin, Neha Gupta, Roman Soletskyi, Sagar Vaze, Teven~Le Scao, Yihan Wang, Adam Yang, Alexander~H. Liu, Alexandre Sablayrolles, Amélie Héliou, Amélie Martin, Andy Ehrenberg, Anmol Agarwal, Antoine Roux, Arthur Darcet, Arthur Mensch, Baptiste Bout, Baptiste Rozière, Baudouin~De Monicault, Chris Bamford, Christian Wallenwein, Christophe Renaudin, Clémence Lanfranchi, Darius Dabert, Devon Mizelle, Diego de~las Casas, Elliot Chane-Sane, Emilien Fugier, Emma~Bou Hanna, Gauthier Delerce, Gauthier Guinet, Georgii Novikov, Guillaume Martin, Himanshu Jaju, Jan Ludziejewski, Jean-Hadrien Chabran, Jean-Malo Delignon, Joachim Studnia, Jonas Amar, Josselin~Somerville Roberts, Julien Denize, Karan Saxena, Kush Jain, Lingxiao Zhao, Louis Martin, Luyu Gao, Lélio~Renard Lavaud, Marie
  Pellat, Mathilde Guillaumin, Mathis Felardos, Maximilian Augustin, Mickaël Seznec, Nikhil Raghuraman, Olivier Duchenne, Patricia Wang, Patrick von Platen, Patryk Saffer, Paul Jacob, Paul Wambergue, Paula Kurylowicz, Pavankumar~Reddy Muddireddy, Philomène Chagniot, Pierre Stock, Pravesh Agrawal, Romain Sauvestre, Rémi Delacourt, Sanchit Gandhi, Sandeep Subramanian, Shashwat Dalal, Siddharth Gandhi, Soham Ghosh, Srijan Mishra, Sumukh Aithal, Szymon Antoniak, Thibault Schueller, Thibaut Lavril, Thomas Robert, Thomas Wang, Timothée Lacroix, Valeriia Nemychnikova, Victor Paltz, Virgile Richard, Wen-Ding Li, William Marshall, Xuanyu Zhang, and Yunhao Tang.
\newblock Magistral, 2025.
\newblock URL \url{https://arxiv.org/abs/2506.10910}.

\bibitem[Ho et~al.(2020)Ho, Jain, and Abbeel]{ho2020denoising}
Jonathan Ho, Ajay Jain, and Pieter Abbeel.
\newblock Denoising diffusion probabilistic models.
\newblock \emph{Advances in neural information processing systems}, 2020.

\bibitem[Song et~al.(2022)Song, Meng, and Ermon]{song2022denoisingdiffusionimplicitmodels}
Jiaming Song, Chenlin Meng, and Stefano Ermon.
\newblock Denoising diffusion implicit models, 2022.
\newblock URL \url{https://arxiv.org/abs/2010.02502}.

\bibitem[Rombach et~al.(2022)Rombach, Blattmann, Lorenz, Esser, and Ommer]{rombach2022highresolutionimagesynthesislatent}
Robin Rombach, Andreas Blattmann, Dominik Lorenz, Patrick Esser, and Björn Ommer.
\newblock High-resolution image synthesis with latent diffusion models, 2022.
\newblock URL \url{https://arxiv.org/abs/2112.10752}.

\bibitem[Song and Ermon(2020)]{song2020generativemodelingestimatinggradients}
Yang Song and Stefano Ermon.
\newblock Generative modeling by estimating gradients of the data distribution, 2020.
\newblock URL \url{https://arxiv.org/abs/1907.05600}.

\bibitem[Ho et~al.(2022{\natexlab{b}})Ho, Salimans, Gritsenko, Chan, Norouzi, and Fleet]{ho2022videodiffusionmodels}
Jonathan Ho, Tim Salimans, Alexey Gritsenko, William Chan, Mohammad Norouzi, and David~J. Fleet.
\newblock Video diffusion models, 2022{\natexlab{b}}.
\newblock URL \url{https://arxiv.org/abs/2204.03458}.

\bibitem[Singer et~al.(2022)Singer, Polyak, Hayes, Yin, An, Zhang, Hu, Yang, Ashual, Gafni, Parikh, Gupta, and Taigman]{singer2022makeavideotexttovideogenerationtextvideo}
Uriel Singer, Adam Polyak, Thomas Hayes, Xi~Yin, Jie An, Songyang Zhang, Qiyuan Hu, Harry Yang, Oron Ashual, Oran Gafni, Devi Parikh, Sonal Gupta, and Yaniv Taigman.
\newblock Make-a-video: Text-to-video generation without text-video data, 2022.
\newblock URL \url{https://arxiv.org/abs/2209.14792}.

\bibitem[Ho et~al.(2022{\natexlab{c}})Ho, Chan, Saharia, Whang, Gao, Gritsenko, Kingma, Poole, Norouzi, Fleet, and Salimans]{ho2022imagenvideohighdefinition}
Jonathan Ho, William Chan, Chitwan Saharia, Jay Whang, Ruiqi Gao, Alexey Gritsenko, Diederik~P. Kingma, Ben Poole, Mohammad Norouzi, David~J. Fleet, and Tim Salimans.
\newblock Imagen video: High definition video generation with diffusion models, 2022{\natexlab{c}}.
\newblock URL \url{https://arxiv.org/abs/2210.02303}.

\bibitem[Popov et~al.(2021)Popov, Vovk, Gogoryan, Sadekova, and Kudinov]{popov2021gradttsdiffusionprobabilisticmodel}
Vadim Popov, Ivan Vovk, Vladimir Gogoryan, Tasnima Sadekova, and Mikhail Kudinov.
\newblock Grad-tts: A diffusion probabilistic model for text-to-speech, 2021.
\newblock URL \url{https://arxiv.org/abs/2105.06337}.

\bibitem[Chen et~al.(2021{\natexlab{b}})Chen, Zhang, Zen, Weiss, Norouzi, Dehak, and Chan]{chen2021wavegrad2iterativerefinement}
Nanxin Chen, Yu~Zhang, Heiga Zen, Ron~J. Weiss, Mohammad Norouzi, Najim Dehak, and William Chan.
\newblock Wavegrad 2: Iterative refinement for text-to-speech synthesis, 2021{\natexlab{b}}.
\newblock URL \url{https://arxiv.org/abs/2106.09660}.

\bibitem[Black et~al.(2024)Black, Brown, Driess, Esmail, Equi, Finn, Fusai, Groom, Hausman, Ichter, Jakubczak, Jones, Ke, Levine, Li-Bell, Mothukuri, Nair, Pertsch, Shi, Tanner, Vuong, Walling, Wang, and Zhilinsky]{black2024pi0visionlanguageactionflowmodel}
Kevin Black, Noah Brown, Danny Driess, Adnan Esmail, Michael Equi, Chelsea Finn, Niccolo Fusai, Lachy Groom, Karol Hausman, Brian Ichter, Szymon Jakubczak, Tim Jones, Liyiming Ke, Sergey Levine, Adrian Li-Bell, Mohith Mothukuri, Suraj Nair, Karl Pertsch, Lucy~Xiaoyang Shi, James Tanner, Quan Vuong, Anna Walling, Haohuan Wang, and Ury Zhilinsky.
\newblock $\pi_0$: A vision-language-action flow model for general robot control, 2024.
\newblock URL \url{https://arxiv.org/abs/2410.24164}.

\bibitem[NVIDIA et~al.(2025)NVIDIA, :, Bjorck, Castañeda, Cherniadev, Da, Ding, Fan, Fang, Fox, Hu, Huang, Jang, Jiang, Kautz, Kundalia, Lao, Li, Lin, Lin, Liu, Llontop, Magne, Mandlekar, Narayan, Nasiriany, Reed, Tan, Wang, Wang, Wang, Wang, Xiang, Xie, Xu, Xu, Ye, Yu, Zhang, Zhang, Zhao, Zheng, and Zhu]{nvidia2025gr00tn1openfoundation}
NVIDIA, :, Johan Bjorck, Fernando Castañeda, Nikita Cherniadev, Xingye Da, Runyu Ding, Linxi~"Jim" Fan, Yu~Fang, Dieter Fox, Fengyuan Hu, Spencer Huang, Joel Jang, Zhenyu Jiang, Jan Kautz, Kaushil Kundalia, Lawrence Lao, Zhiqi Li, Zongyu Lin, Kevin Lin, Guilin Liu, Edith Llontop, Loic Magne, Ajay Mandlekar, Avnish Narayan, Soroush Nasiriany, Scott Reed, You~Liang Tan, Guanzhi Wang, Zu~Wang, Jing Wang, Qi~Wang, Jiannan Xiang, Yuqi Xie, Yinzhen Xu, Zhenjia Xu, Seonghyeon Ye, Zhiding Yu, Ao~Zhang, Hao Zhang, Yizhou Zhao, Ruijie Zheng, and Yuke Zhu.
\newblock Gr00t n1: An open foundation model for generalist humanoid robots, 2025.
\newblock URL \url{https://arxiv.org/abs/2503.14734}.

\bibitem[Skreta et~al.(2025)Skreta, Atanackovic, Bose, Tong, and Neklyudov]{skreta2025superpositiondiffusionmodelsusing}
Marta Skreta, Lazar Atanackovic, Avishek~Joey Bose, Alexander Tong, and Kirill Neklyudov.
\newblock The superposition of diffusion models using the it\^o density estimator, 2025.
\newblock URL \url{https://arxiv.org/abs/2412.17762}.

\bibitem[Chi et~al.(2024{\natexlab{b}})Chi, Xu, Feng, Cousineau, Du, Burchfiel, Tedrake, and Song]{chi2023diffusion}
Cheng Chi, Zhenjia Xu, Siyuan Feng, Eric Cousineau, Yilun Du, Benjamin Burchfiel, Russ Tedrake, and Shuran Song.
\newblock Diffusion policy: Visuomotor policy learning via action diffusion.
\newblock \emph{The International Journal of Robotics Research}, 2024{\natexlab{b}}.

\bibitem[Ajay et~al.(2023)Ajay, Du, Gupta, Tenenbaum, Jaakkola, and Agrawal]{ajay2022conditional}
Anurag Ajay, Yilun Du, Abhi Gupta, Joshua~B. Tenenbaum, Tommi~S. Jaakkola, and Pulkit Agrawal.
\newblock Is conditional generative modeling all you need for decision making?
\newblock In \emph{The Eleventh International Conference on Learning Representations}, 2023.

\bibitem[Janner et~al.(2022)Janner, Du, Tenenbaum, and Levine]{janner2022planning}
Michael Janner, Yilun Du, Joshua~B Tenenbaum, and Sergey Levine.
\newblock Planning with diffusion for flexible behavior synthesis.
\newblock \emph{arXiv preprint arXiv:2205.09991}, 2022.

\bibitem[Lee et~al.(2023)Lee, Liu, Ryu, Watkins, Du, Boutilier, Abbeel, Ghavamzadeh, and Gu]{lee2023aligning}
Kimin Lee, Hao Liu, Moonkyung Ryu, Olivia Watkins, Yuqing Du, Craig Boutilier, Pieter Abbeel, Mohammad Ghavamzadeh, and Shixiang~Shane Gu.
\newblock Aligning text-to-image models using human feedback.
\newblock \emph{arXiv preprint arXiv:2302.12192}, 2023.

\bibitem[Black et~al.(2023)Black, Janner, Du, Kostrikov, and Levine]{black2023training}
Kevin Black, Michael Janner, Yilun Du, Ilya Kostrikov, and Sergey Levine.
\newblock Training diffusion models with reinforcement learning.
\newblock \emph{arXiv preprint arXiv:2305.13301}, 2023.

\bibitem[Liu et~al.(2025)Liu, Liu, Liang, Li, Liu, Wang, Wan, Zhang, and Ouyang]{liu2025flow}
Jie Liu, Gongye Liu, Jiajun Liang, Yangguang Li, Jiaheng Liu, Xintao Wang, Pengfei Wan, Di~Zhang, and Wanli Ouyang.
\newblock Flow-grpo: Training flow matching models via online rl.
\newblock \emph{arXiv preprint arXiv:2505.05470}, 2025.

\bibitem[Psenka et~al.(2023)Psenka, Escontrela, Abbeel, and Ma]{psenka2023learning}
Michael Psenka, Alejandro Escontrela, Pieter Abbeel, and Yi~Ma.
\newblock Learning a diffusion model policy from rewards via q-score matching.
\newblock \emph{arXiv preprint arXiv:2312.11752}, 2023.

\bibitem[Seo et~al.(2025)Seo, Sferrazza, Geng, Nauman, Yin, and Abbeel]{seo2025fasttd3simplefastcapable}
Younggyo Seo, Carmelo Sferrazza, Haoran Geng, Michal Nauman, Zhao-Heng Yin, and Pieter Abbeel.
\newblock Fasttd3: Simple, fast, and capable reinforcement learning for humanoid control, 2025.
\newblock URL \url{https://arxiv.org/abs/2505.22642}.

\bibitem[Fujimoto et~al.(2018)Fujimoto, van Hoof, and Meger]{fujimoto2018addressingfunctionapproximationerror}
Scott Fujimoto, Herke van Hoof, and David Meger.
\newblock Addressing function approximation error in actor-critic methods, 2018.
\newblock URL \url{https://arxiv.org/abs/1802.09477}.

\bibitem[Ren et~al.(2024)Ren, Lidard, Ankile, Simeonov, Agrawal, Majumdar, Burchfiel, Dai, and Simchowitz]{ren2024diffusion}
Allen~Z Ren, Justin Lidard, Lars~L Ankile, Anthony Simeonov, Pulkit Agrawal, Anirudha Majumdar, Benjamin Burchfiel, Hongkai Dai, and Max Simchowitz.
\newblock Diffusion policy policy optimization.
\newblock \emph{arXiv preprint arXiv:2409.00588}, 2024.

\bibitem[Schulman et~al.(2015{\natexlab{b}})Schulman, Moritz, Levine, Jordan, and Abbeel]{schulman2015high}
John Schulman, Philipp Moritz, Sergey Levine, Michael Jordan, and Pieter Abbeel.
\newblock High-dimensional continuous control using generalized advantage estimation.
\newblock \emph{arXiv preprint arXiv:1506.02438}, 2015{\natexlab{b}}.

\bibitem[Karras et~al.(2022)Karras, Aittala, Aila, and Laine]{karras2022elucidating}
Tero Karras, Miika Aittala, Timo Aila, and Samuli Laine.
\newblock Elucidating the design space of diffusion-based generative models.
\newblock \emph{Advances in neural information processing systems}, 35:\penalty0 26565--26577, 2022.

\bibitem[Kingma et~al.(2023)Kingma, Salimans, Poole, and Ho]{kingma2023variationaldiffusionmodels}
Diederik~P. Kingma, Tim Salimans, Ben Poole, and Jonathan Ho.
\newblock Variational diffusion models, 2023.
\newblock URL \url{https://arxiv.org/abs/2107.00630}.

\bibitem[Kingma and Gao(2023)]{kingma2023understandingdiffusionobjectiveselbo}
Diederik~P. Kingma and Ruiqi Gao.
\newblock Understanding diffusion objectives as the elbo with simple data augmentation, 2023.
\newblock URL \url{https://arxiv.org/abs/2303.00848}.

\bibitem[Brockman et~al.(2016)Brockman, Cheung, Pettersson, Schneider, Schulman, Tang, and Zaremba]{brockman2016gym}
Greg Brockman, Vicki Cheung, Ludwig Pettersson, Jonas Schneider, John Schulman, Jie Tang, and Wojciech Zaremba.
\newblock Openai gym, 2016.

\bibitem[Towers et~al.(2024)Towers, Kwiatkowski, Terry, Balis, De~Cola, Deleu, Goul{\~a}o, Kallinteris, Krimmel, KG, et~al.]{towers2024gymnasium}
Mark Towers, Ariel Kwiatkowski, Jordan Terry, John~U Balis, Gianluca De~Cola, Tristan Deleu, Manuel Goul{\~a}o, Andreas Kallinteris, Markus Krimmel, Arjun KG, et~al.
\newblock Gymnasium: A standard interface for reinforcement learning environments.
\newblock \emph{arXiv preprint arXiv:2407.17032}, 2024.

\bibitem[Todorov et~al.(2012)Todorov, Erez, and Tassa]{todorov2012mujoco}
Emanuel Todorov, Tom Erez, and Yuval Tassa.
\newblock Mujoco: A physics engine for model-based control.
\newblock In \emph{2012 IEEE/RSJ international conference on intelligent robots and systems}, 2012.

\bibitem[Makoviychuk et~al.(2021)Makoviychuk, Wawrzyniak, Guo, Lu, Storey, Macklin, Hoeller, Rudin, Allshire, Handa, et~al.]{makoviychuk2021isaac}
Viktor Makoviychuk, Lukasz Wawrzyniak, Yunrong Guo, Michelle Lu, Kier Storey, Miles Macklin, David Hoeller, Nikita Rudin, Arthur Allshire, Ankur Handa, et~al.
\newblock Isaac gym: High performance gpu-based physics simulation for robot learning.
\newblock \emph{arXiv preprint arXiv:2108.10470}, 2021.

\bibitem[Yu(2020)]{yu2020ppo_for_beginners}
Eric~Yang Yu.
\newblock Ppo-for-beginners: A simple, well-styled ppo implementation in pytorch.
\newblock \url{https://github.com/ericyangyu/PPO-for-Beginners}, 2020.
\newblock GitHub repository.

\bibitem[Tassa et~al.(2018)Tassa, Doron, Muldal, Erez, Li, Casas, Budden, Abdolmaleki, Merel, Lefrancq, et~al.]{tassa2018deepmind}
Yuval Tassa, Yotam Doron, Alistair Muldal, Tom Erez, Yazhe Li, Diego de~Las Casas, David Budden, Abbas Abdolmaleki, Josh Merel, Andrew Lefrancq, et~al.
\newblock Deepmind control suite.
\newblock \emph{arXiv preprint arXiv:1801.00690}, 2018.

\bibitem[Tunyasuvunakool et~al.(2020)Tunyasuvunakool, Muldal, Doron, Liu, Bohez, Merel, Erez, Lillicrap, Heess, and Tassa]{tunyasuvunakool2020dm_control}
Saran Tunyasuvunakool, Alistair Muldal, Yotam Doron, Siqi Liu, Steven Bohez, Josh Merel, Tom Erez, Timothy Lillicrap, Nicolas Heess, and Yuval Tassa.
\newblock dm\_control: Software and tasks for continuous control.
\newblock \emph{Software Impacts}, 6:\penalty0 100022, 2020.

\bibitem[Freeman et~al.(2021)Freeman, Frey, Raichuk, Girgin, Mordatch, and Bachem]{freeman2021brax}
C~Daniel Freeman, Erik Frey, Anton Raichuk, Sertan Girgin, Igor Mordatch, and Olivier Bachem.
\newblock Brax--a differentiable physics engine for large scale rigid body simulation.
\newblock \emph{arXiv preprint arXiv:2106.13281}, 2021.

\bibitem[Kingma(2014)]{kingma2014adam}
Diederik~P Kingma.
\newblock Adam: A method for stochastic optimization.
\newblock \emph{arXiv preprint arXiv:1412.6980}, 2014.

\bibitem[Luo et~al.(2023{\natexlab{a}})Luo, Cao, Kitani, Xu, et~al.]{luo2023perpetual}
Zhengyi Luo, Jinkun Cao, Kris Kitani, Weipeng Xu, et~al.
\newblock Perpetual humanoid control for real-time simulated avatars.
\newblock In \emph{Proceedings of the IEEE/CVF International Conference on Computer Vision}, pages 10895--10904, 2023{\natexlab{a}}.

\bibitem[Peng et~al.(2018)Peng, Abbeel, Levine, and Van~de Panne]{peng2018deepmimic}
Xue~Bin Peng, Pieter Abbeel, Sergey Levine, and Michiel Van~de Panne.
\newblock Deepmimic: Example-guided deep reinforcement learning of physics-based character skills.
\newblock \emph{ACM Transactions On Graphics (TOG)}, 37\penalty0 (4):\penalty0 1--14, 2018.

\bibitem[Mahmood et~al.(2019)Mahmood, Ghorbani, Troje, Pons-Moll, and Black]{mahmood2019amass}
Naureen Mahmood, Nima Ghorbani, Nikolaus~F Troje, Gerard Pons-Moll, and Michael~J Black.
\newblock Amass: Archive of motion capture as surface shapes.
\newblock In \emph{Proceedings of the IEEE/CVF international conference on computer vision}, pages 5442--5451, 2019.

\bibitem[Tessler et~al.(2024)Tessler, Guo, Nabati, Chechik, and Peng]{tessler2024maskedmimic}
Chen Tessler, Yunrong Guo, Ofir Nabati, Gal Chechik, and Xue~Bin Peng.
\newblock Maskedmimic: Unified physics-based character control through masked motion inpainting.
\newblock \emph{ACM Transactions on Graphics (TOG)}, 43\penalty0 (6):\penalty0 1--21, 2024.

\bibitem[Li et~al.(2025)Li, Lin, Cui, Liu, Liang, Zhu, and Huang]{li2025clone}
Yixuan Li, Yutang Lin, Jieming Cui, Tengyu Liu, Wei Liang, Yixin Zhu, and Siyuan Huang.
\newblock Clone: Closed-loop whole-body humanoid teleoperation for long-horizon tasks.
\newblock \emph{arXiv preprint arXiv:2506.08931}, 2025.

\bibitem[Luo et~al.(2023{\natexlab{b}})Luo, Cao, Merel, Winkler, Huang, Kitani, and Xu]{luo2023universal}
Zhengyi Luo, Jinkun Cao, Josh Merel, Alexander Winkler, Jing Huang, Kris Kitani, and Weipeng Xu.
\newblock Universal humanoid motion representations for physics-based control.
\newblock \emph{arXiv preprint arXiv:2310.04582}, 2023{\natexlab{b}}.

\bibitem[Luo et~al.(2024)Luo, Cao, Christen, Winkler, Kitani, and Xu]{luo2024omnigrasp}
Zhengyi Luo, Jinkun Cao, Sammy Christen, Alexander Winkler, Kris Kitani, and Weipeng Xu.
\newblock Omnigrasp: Grasping diverse objects with simulated humanoids.
\newblock In \emph{Advances in Neural Information Processing Systems}, volume~37, pages 2161--2184, 2024.

\bibitem[Shumailov et~al.(2024)Shumailov, Shumaylov, Zhao, Papernot, Anderson, and Gal]{shumailov2024ai}
Ilia Shumailov, Zakhar Shumaylov, Yiren Zhao, Nicolas Papernot, Ross Anderson, and Yarin Gal.
\newblock Ai models collapse when trained on recursively generated data.
\newblock \emph{Nature}, 631\penalty0 (8022):\penalty0 755--759, 2024.

\bibitem[Shumailov et~al.(2023)Shumailov, Shumaylov, Zhao, Gal, Papernot, and Anderson]{shumailov2023curse}
Ilia Shumailov, Zakhar Shumaylov, Yiren Zhao, Yarin Gal, Nicolas Papernot, and Ross Anderson.
\newblock The curse of recursion: Training on generated data makes models forget.
\newblock \emph{arXiv preprint arXiv:2305.17493}, 2023.

\bibitem[Alemohammad et~al.(2024)Alemohammad, Casco-Rodriguez, Luzi, Humayun, Babaei, LeJeune, Siahkoohi, and Baraniuk]{alemohammad2024self}
Sina Alemohammad, Josue Casco-Rodriguez, Lorenzo Luzi, Ahmed~Imtiaz Humayun, Hossein Babaei, Daniel LeJeune, Ali Siahkoohi, and Richard~G Baraniuk.
\newblock Self-consuming generative models go mad.
\newblock International Conference on Learning Representations (ICLR), 2024.

\bibitem[Salimans and Ho(2022)]{salimans2022progressive}
Tim Salimans and Jonathan Ho.
\newblock Progressive distillation for fast sampling of diffusion models.
\newblock \emph{arXiv preprint arXiv:2202.00512}, 2022.

\bibitem[Ho and Salimans(2022)]{ho2022classifierfreediffusionguidance}
Jonathan Ho and Tim Salimans.
\newblock Classifier-free diffusion guidance, 2022.
\newblock URL \url{https://arxiv.org/abs/2207.12598}.

\end{thebibliography}

\renewcommand{\thetable}{A.\arabic{table}}
\setcounter{table}{0}  %
\renewcommand{\thesection}{A.\arabic{section}}
\setcounter{section}{0}  %
\renewcommand{\thefigure}{A.\arabic{figure}}
\setcounter{figure}{0}  %

\onecolumn{
\centering
\Large
\textbf{Flow Matching Policy Gradients}\\
\vspace{0.3em}Supplementary Material \\
\vspace{1.0em}
}

In this supplementary material, we discuss the deferred proofs of technical results, elaborate on the details of our experiments, and present additional visual results for the grid world, humanoid control, and image finetuning experiments.

\section{FPO Derivation}

The mathematical details presented in this section provide expanded derivations and additional context for the theoretical results outlined in Section 3 of the main text. Specifically, we elaborate on the connection between the conditional flow matching objective and the evidence lower bound (ELBO) first mentioned in Section 3.4, and provide complete derivations for the FPO ratio introduced in Section 3.3. These details are included for completeness and to situate our work within the theoretical framework established by Kingma et al. \cite{kingma2023understandingdiffusionobjectiveselbo}, but are not necessary for understanding the core FPO algorithm or implementing it in practice.

First, we detail the different popular loss weightings used when training flow matching models laid out by Kingma et al. \cite{kingma2023understandingdiffusionobjectiveselbo}. These weightings, denoted as $w(\lambda_t)$, determine how losses at different noise levels contribute to the overall objective and lead to different theoretical interpretations of Flow Policy Optimization.

Then, we show the more general result, which is that FPO optimizes the advantage-weighted expected ELBO of the noise-perturbed data. Specifically, for any monotonic weighting function (including Optimal Transport CFM schedules~\cite{lipman2023flowmatchinggenerativemodeling}), we can express the weighted loss as:
\begin{align}
    \weightedloss(a_t) = -\mathbb{E}_{p_w(\tau),q(\att|a_t)}[\text{ELBO}_\tau(\att)] + c_1,
\end{align}
where $p_w(\tau)$ is the distribution over timesteps induced by the weighting function, and $\text{ELBO}_\tau(\att)$ is the evidence lower bound at noise level $\tau$ for the perturbed action $\att$.

This means that FPO increases the likelihood of high-reward samples and the intermediate noisy samples $\att$ from the sample path. By weighting this objective with advantages $\hat{A}_\tau$, we guide the policy to direct probability flow toward action neighborhoods that produce higher reward.

For diffusion schedules with uniform weighting $w(\lambda_\tau) = 1$, we show a somewhat stronger theoretical result. In this special case, the weighted loss directly corresponds to maximizing the ELBO of clean actions:
\begin{align}
-\text{ELBO}(a_t) = \frac{1}{2}\mathbb{E}_{\tau\sim U(0,1),\epsilon\sim\mathcal{N}(0,I)}\left[-\frac{d\lambda}{d\tau} \cdot \|\hat{\epsilon}_\theta(\att; \lambda_\tau) - \epsilon\|^2_2\right] + c_2,
\end{align}
which is a more direct connection to maximum likelihood estimation.

\subsection{Loss Weighting Choices}

Most popular instantiations of flow-based and diffusion models can be reparameterized in the weighted loss scheme proposed by Kingma et al.~\cite{kingma2023understandingdiffusionobjectiveselbo}. This unified framework expresses each version as an instance of a weighted denoising loss:
\begin{align}
    \weightedloss(x) = \frac{1}{2}\mathbb{E}_{\tau\sim U(0,1),\epsilon\sim\mathcal{N}(0,I)}[w(\lambda_\tau) \cdot -\frac{d\lambda}{d\tau} \cdot \|\hat{\epsilon}_\theta(\att; \lambda_\tau) - \epsilon\|^2_2],
\end{align}
where $w(\lambda_\tau)$ is a time-dependent function that determines the relative importance of different noise levels.

For those with a loss weight that varies monotonically with noise timestep $\tau$, the aforementioned relationship between the weighted loss and expected ELBO holds. Specifically, when $w(\lambda_\tau)$ is monotonically increasing with $\tau$, Kingma et al. prove:
\begin{align}
\weightedloss(a_t) = -\mathbb{E}_{p_w(\tau),q(\att|a_t)}[\text{ELBO}_\tau(\att)] + c_1,
\end{align}
where $c_1$ is a constant, and does not vary with model parameters.

These monotonic weightings include several popular schedules: (1) standard diffusion with uniform weighting $w(\lambda_\tau) = 1$~\cite{ho2020denoising}, (2) optimal transport linear interpolation schedule~\cite{lipman2023flowmatchinggenerativemodeling}, which yields $w(\lambda_\tau) = e^{-\lambda/2}$, and (3) velocity prediction (v-prediction) with cosine schedule~\cite{salimans2022progressive}, which also yields $w(\lambda_\tau) = e^{-\lambda/2}$.

\subsection{Flow Matching as Expected ELBO Optimization}
To derive FPO in the more general flow matching case, we begin with the standard policy gradient objective, but replace direct likelihood maximization with maximization of the ELBO for noise-perturbed data:

\begin{align}
\max_\theta \mathbb{E}_{a_t \sim \pi_\theta(a_t|o_t)}\left[\mathbb{E}_{p_w(\tau),q(\att|a_t)}[\text{ELBO}_\tau(\att)] \cdot \hat{A}_t\right],
\end{align}

where $t$ is temporal rollout time and $\tau$ is diffusion/flow noise timestep.

This formulation directly leverages the result from Kingma et al. \cite{kingma2023understandingdiffusionobjectiveselbo} that for monotonic weightings, the weighted denoising loss equals the negative expected ELBO of noise-perturbed data plus a constant:
\begin{align}
\weightedloss(a_t) = -\mathbb{E}_{p_w(\tau),q(\att|a_t)}[\text{ELBO}_\tau(\att)] + c_1.
\end{align}

To apply this within a trust region approach similar to PPO, we need to define a ratio between the current and old policies. Since we are working with expected ELBOs, the appropriate ratio becomes:

\begin{align}
r^{\text{FPO}}(\theta) = \frac{\exp(\mathbb{E}_{p_w(\tau),q(\att|a_t)}[\text{ELBO}_\tau(\att)]_\theta)}{\exp(\mathbb{E}_{p_w(\tau),q(\att|a_t)}[\text{ELBO}_\tau(\att)]_{\theta,\text{old}})}
\end{align}

This ratio represents the relative likelihood of actions and their noisy versions under the current policy compared to the old policy.

It is important to note that the constant $c_1$ in the ELBO equivalence depends only on the noise schedule endpoints $\lambda_{min}$ and $\lambda_{max}$, the data distribution, and the forward process, but not on the model parameter $\theta$. This is critical for our derivation. It ensures that within a single trust region data collection and training episode, this constant remains identical between the old policy $\theta_{old}$ and the updated policy $\theta$. Consequently, when forming the ratio $r^{\text{FPO}}(\theta)$, these constants cancel out:

\begin{align}
r^{\text{FPO}}(\theta) = \frac{\exp(\mathbb{E}_{p_w(\tau),q(\att|a_t)}[\text{ELBO}_\tau(\att)]_\theta+c_1)}{\exp(\mathbb{E}_{p_w(\tau),q(\att|a_t)}[\text{ELBO}_\tau(\att)]_{\theta,\text{old}}+c_1)}
= \frac{\exp(\mathbb{E}_{p_w(\tau),q(\att|a_t)}[\text{ELBO}_\tau(\att)]_\theta)}{\exp(\mathbb{E}_{p_w(\tau),q(\att|a_t)}[\text{ELBO}_\tau(\att)]_{\theta,\text{old}})}
\end{align}

We estimate this ratio through Monte Carlo sampling of timesteps $\tau$ and noise $\epsilon$:
\begin{align}
\hat{r}^{\text{FPO}}(\tau,\epsilon) = \exp(-\ell_\theta(\tau,\epsilon) + \ell_{\theta,\text{old}}(\tau,\epsilon)),
\end{align}
where $\ell_\theta(\tau,\epsilon) = \frac{1}{2}[-\dot{\lambda}(\tau)] \|\hat{\epsilon}_\theta(\att; \lambda_\tau) - \epsilon\|^2$ is the reparameterized conditional flow matching loss for a single draw of random variables $\epsilon$ and $\tau$.

As discussed in the main text, $\hat{r}^{\text{FPO}}$ overestimates the scale but unbiasedly estimates the direction of the gradient. We can reduce or eliminate the scale bias by drawing more samples of $\tau$ and $\epsilon$.

\subsection{FPO with Diffusion Schedules}

For the special case of standard diffusion schedules with uniform weighting $w(\lambda_t) = 1$, we can derive a stronger theoretical result connecting our optimization objective directly to the ELBO of clean (non-noised) data.

As shown by Kingma et al. \cite{kingma2023understandingdiffusionobjectiveselbo}, when using uniform weighting, the weighted loss directly corresponds to the negative ELBO of the clean data plus a constant:
\begin{align}
-\text{ELBO}(a_t) = \frac{1}{2}\mathbb{E}_{\tau\sim U(0,1),\epsilon\sim\mathcal{N}(0,I)}\left[-\frac{d\lambda}{d\tau} \cdot \|\hat{\epsilon}_\theta(\att; \lambda_\tau) - \epsilon\|^2_2\right] + c_2,
\end{align}

where $c_2$ is a different constant than $c_1$ that also does not depend on model parameter $\theta$.

This means that minimizing the unweighted loss ($w(\lambda_\tau) = 1$) is equivalent to maximizing the ELBO of the clean action $a_t$, providing a more direct connection to traditional maximum likelihood estimation.

In the context of FPO, we can therefore express our advantage-weighted objective as:
\begin{align}
\max_\theta \mathbb{E}_{a_t \sim \pi_\theta(a_t|o_t)}\left[\text{ELBO}_\theta(a_t) \cdot \hat{A}_t\right]
\end{align}
In this case, the objective direct increases a lower bound of the log-likelihood of clean actions $a_t$ weighted by their advantages, rather than over noise-perturbed actions.  %

The FPO ratio in this case becomes:
\begin{align}
r^{\text{FPO}}(\theta) = \frac{\exp(\text{ELBO}_\theta(a_t))}{\exp(\text{ELBO}_{\theta,\text{old}}(a_t))}
\end{align}

This specific case highlights the close relationship between FPO and traditional maximum likelihood methods common for PPO~\cite{schulman2017proximal}. FPO still retains the computational advantages of avoiding explicit likelihood computations.

As in the general case, our Monte Carlo estimator exhibits upward bias of gradient scale. We can use the same PPO clipping mechanism to control the magnitude of parameter changes.

\subsection{Advantage-Weighed Flow Matching Discussion}

Advantage estimates are typically zero-centered to reduce variance in estimating the policy gradient. Flow matching, however, learns probability flows which must be nonnegative by construction. Since advantages function as loss weights in this context, they should remain positive for mathematical consistency. A constant shift does not affect policy gradient optimization, which follows from the same baseline-invariance property that justifies using advantages in the first place. We find that both processed and unprocessed advantages work empirically.

\begin{figure}[t!]
    \centering
    \includegraphics[width=\linewidth,clip, trim=1cm 6.5cm 1cm 6.5cm]{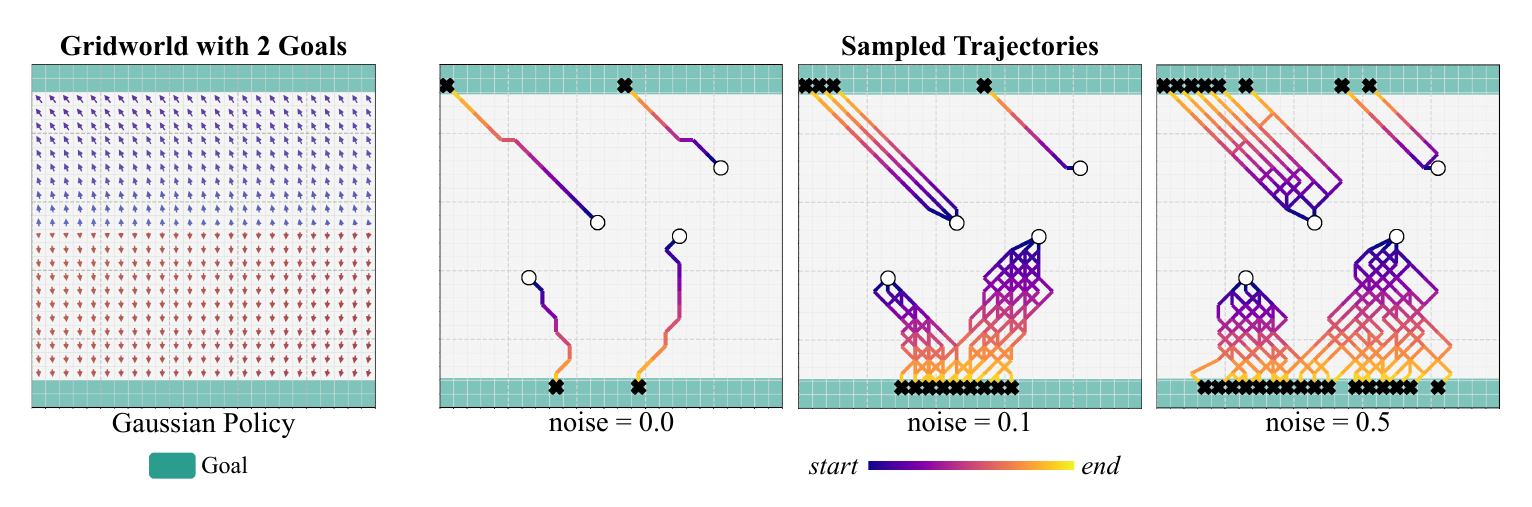}
    \caption{\textbf{GridWorld with Gaussian Policy.} Left) $25\times 25$ GridWorld with green goal cells. Each arrow shows an action predicted by the Gaussian policy. Right) Four rollouts under test-time noise perturbations ($\sigma=0.0$, $0.1$, $0.5$). While the Gaussian policy achieves the goal, its trajectories lack diversity and hit the same goal consistently when given the same initialization point.}
    \label{fig:gridgs}
\end{figure}

\section{GridWorld}
\label{app:gridworld}
Figure~\ref{fig:gridgs} shows results from the Gaussian policy on the same Grid World trained using PPO. While the Gaussian policy can learn optimal behaviors, the trajectories resulting from it are not as diverse as those of the diffusion policy. 
We visualize 4 samples from the Gaussian policy with 0.0, 0.1, and 0.5 random noise perturbations at test time (Fig.~\ref{fig:gridgs}, right). Note that despite being initialized at the midpoint of the environment, all shown positions lead to a \textit{single} goal mode, never both.

\section{MuJoCo Playground}
\label{app:playground_details}

\begin{figure}[!t]
    \centering
    \includegraphics[width=0.99\linewidth]{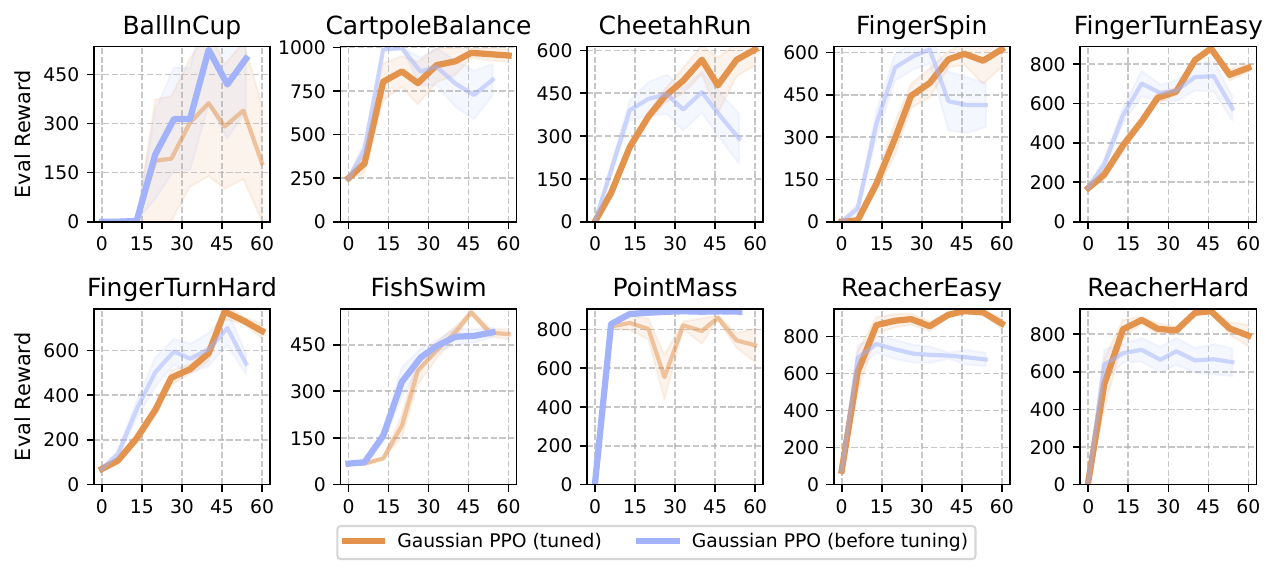}
    \vspace{-.5em}
    \caption{
        \textbf{Gaussian PPO baseline results before and after tuning.}
        We tune clipping epsilon and learning rate to maximize average performance across tasks.
        Results show evaluation reward mean and standard error (y-axis) over 60M environment steps (x-axis).
        We run 5 seeds for each task; the curve with the highest terminal evaluation reward is bolded.
    }
    \label{fig:gaussian_ppo_tune_improvement}
\end{figure}

Table~\ref{tab:mujoco_ppo_hyperparams} shows hyperparameters used for PPO training in the MuJoCo Playground environment.
These are imported directly from the configurations provided by MuJoCo Playground~\cite{zakka2025mujoco}, but after sweeping hyperparameters to tune learning rate and clipping coefficients (Table~\ref{tab:gaussian_ppo_mujoco_sweep}).
We visualize improvements from this sweep in Figure~\ref{fig:gaussian_ppo_tune_improvement}.
Our flow matching and diffusion-based policies use the same hyperparameters, but adjust the clipping coefficient, turn off the entropy coefficient, and for DPPO~\cite{ren2024diffusion}, introduce a stochastic sampling variance to account for the change in policy representation.

\begin{table}[!t]
    \centering
    \begin{tabular}{lcccccc}
        \toprule
        \textbf{Learning Rate} & \multicolumn{6}{c}{\textbf{Clipping Epsilon ($\varepsilon^\text{clip}$)}} \\
        \cmidrule(lr){2-7}
         & 0.3 & 0.2 & 0.1 & 0.05 & 0.03 & 0.01 \\
        \midrule
        0.0001 & 589.5 & 648.5 & 646.6 & 608.6 & 500.5 & 458.5 \\
        0.001 & 556.0 & 646.1 & 654.6 & 636.2 & 562.6 & 471.8 \\
        0.003 & 548.9 & 603.1 & 586.4 & 535.7 & 480.8 & 400.8 \\
        0.0003 & 567.0 & 631.8 & \textbf{667.8} & 650.9 & 570.4 & 492.0 \\
        0.0005 & 544.8 & 586.8 & 629.5 & 559.7 & 505.6 & 406.5 \\
        \bottomrule
    \end{tabular}
    \vspace{1.25em}
    \caption{\textbf{Hyperparameter sweep for Gaussian PPO on the subset of Playground tasks that we evaluate on.}
    All quantities are average rewards across 10 tasks, with 5 seeds per task.
    The default configuration in Playground~\cite{zakka2025mujoco} (before tuning) uses learning rate 1e-3 and clipping epsilon 0.3; the tuned variant we use for results in the main paper body sets learning rate to 3e-4 and clipping epsilon to 0.1.
    }
    \label{tab:gaussian_ppo_mujoco_sweep}
\end{table}

\begin{table}[h]
  \centering
  \begin{tabular}{ll}
  \toprule
  \textbf{Hyperparameter} & \textbf{Value} \\
  \midrule
      Discount factor ($\gamma$) & 0.995 (most environments) \\
       & 0.95 (BallInCup, FingerSpin) \\
      GAE $\lambda$ & 0.95 \\
      Value loss coefficient & 0.25 \\
      Entropy coefficient & 0.01 \\
      Reward scaling & 10.0 \\
      Normalize advantage & True \\
      Normalize observations & True \\
      Action repeat & 1\\
      Unroll length & 30 \\
      Batch size & 1024 \\
      Number of minibatches & 32 \\
      Number of updates per batch & 16 \\
      Number of environments & 2048 \\
      Number of evaluations & 10 \\
      Number of timesteps & 60M \\
      Policy network & MLP (4 hidden layers, 32 units) \\
      Value network & MLP (5 hidden layers, 256 units) \\
      Optimizer & Adam \\
  \bottomrule
  \end{tabular}
  \vspace{0.75em}
  \caption{\textbf{PPO hyperparameters imported from MuJoCo playground~\cite{zakka2025mujoco}.}}
  \label{tab:mujoco_ppo_hyperparams}
\end{table}

\section{Humanoid Control}
\label{sec:phc_details}

In Table~\ref{tab:phc_hyperparam}, we report the detailed hyperparameters that we used for training both the Gaussian policy with PPO and the Diffusion policy with FPO in the humanoid control experiment. Note that we use the same set of hyperparameters for both policies. In our project webpage, we also provide videos
showing qualitative comparisons between the Gaussian policy and ours on tracking an under-conditioned reference, and visual results of FPO on different terrains.

\begin{table}[ht]
\centering
\begin{tabular}{ll | ll}
\toprule
\textbf{Hyperparameter} & \textbf{Value} & \textbf{Hyperparameter} & \textbf{Value} \\
\midrule
\multicolumn{4}{c}{\textit{Policy Settings}} \\
\midrule
Hidden size & 512 & Solver step size & 0.1 \\
Action perturbation std & 0.05 & Target KL divergence & None \\
Number of environments & 4096 & Normalize advantage & True \\
\midrule
\multicolumn{4}{c}{\textit{Training Settings}} \\
\midrule
Batch size & 131072 & Minibatch size & 32768 \\
Learning rate & 0.0001 & LR annealing & False \\
LR decay rate & 1.5e-4 & LR decay floor & 0.2 \\
Update epochs & 4 & L2 regularization coef. & 0.0 \\
GAE lambda & 0.2 & Discount factor ($\gamma$) & 0.98 \\
Clipping coefficient & 0.01 & Value function coefficient & 1.2 \\
Clip value loss & True & Value loss clip coefficient & 0.2 \\
Max gradient norm & 10.0 & Entropy coefficient & 0.0 \\
Discriminator coefficient & 5.0 & Bound coefficient & 10.0 \\
\bottomrule
\end{tabular}
\vspace{0.75em}
\caption{\textbf{Policy training hyperparameters for humanoid control.}}
\label{tab:phc_hyperparam}
\end{table}

\section{Image Reward Fine-tuning}
We explore fine-tuning a pre-trained image diffusion model on a non-differentiable task using the JPEG image compression gym proposed in DDPO~\citep{black2023training}. We report this experiment as a negative result for FPO, due to the difficulty of fine-tuning diffusion models on their own output. 
Specifically, we find that repeatedly generating samples from a text-to-image diffusion model and training on them is highly unstable, even with manually-specified uniform advantages. We believe that this is related to classifier-free guidance (CFG)~\cite{ho2022classifierfreediffusionguidance}. CFG is necessary to generate realistic images, however it is sensitive to hyperparameters, where too much or too little guidance introduces artifacts such as blur or oversaturation that do not reflect the original training data. Sometimes these artifacts are not visible to human eyes. These artifacts are further amplified over successive iterations of RL epochs, ultimately dominating the training signal. %

\begin{figure}[t!]
    \centering
    \includegraphics[width=\linewidth]{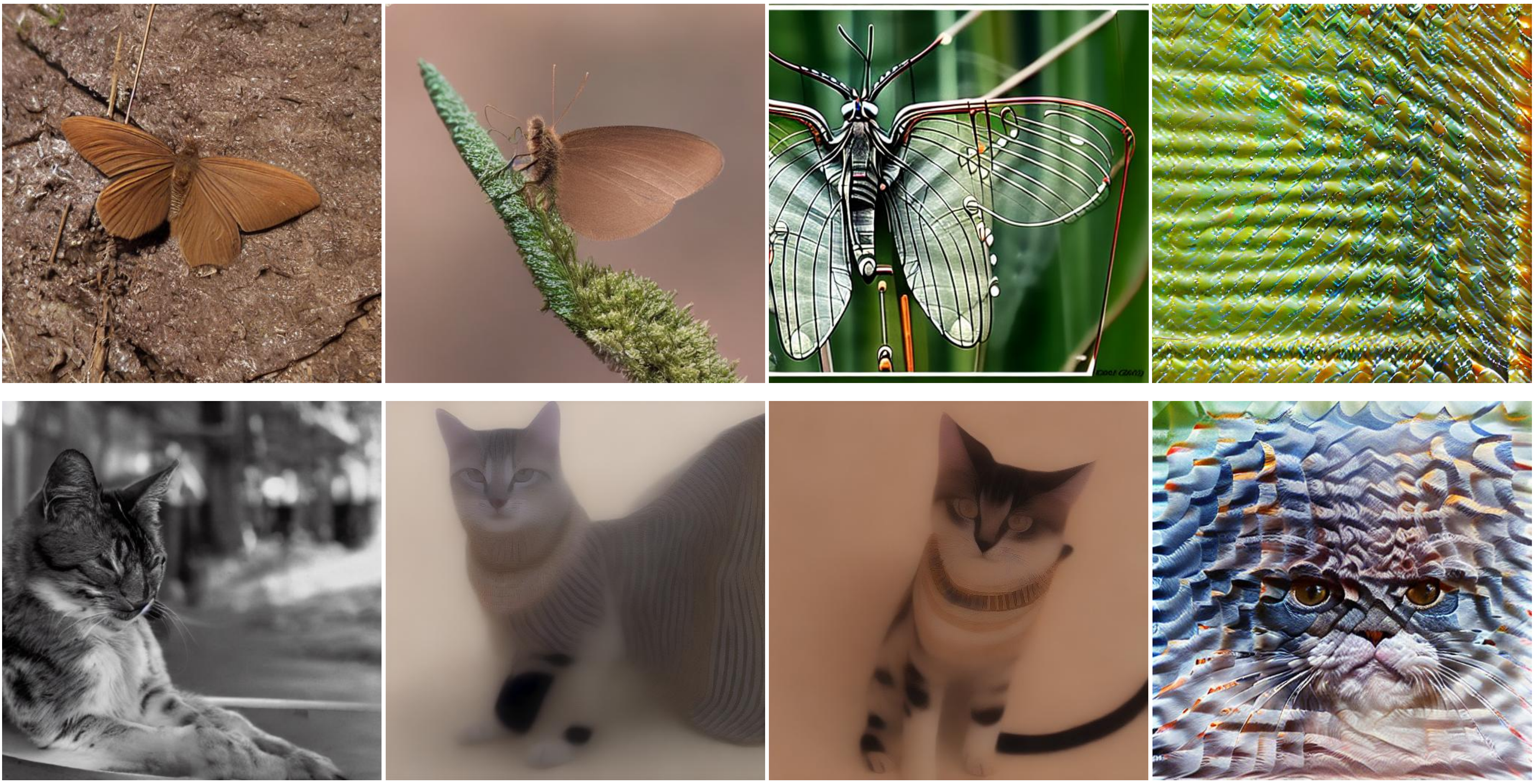}
    \caption{\textbf{Image Generation at Different Training Steps.} We generate images using Stable Diffusion 1.5 finetuned with FPO as training progresses. We manually set all advantages to 1 to eliminate the reward signal and investigate the dynamics of sampling from a text-to-image diffusion model then training on the results in a loop. In the top row, we display images from a training run using a classifier-free guidance (CFG) scale of 4. In the bottom row, we display images from a training run using a CFG scale of 2. Low CFG scales tend to encourage bluriness while high CFG scales encourage saturation and sharp geometric artifacts. Both diverge after a few hundred epochs even with tuned hyperparameters.}
\label{fig:fpo_image_supp}
\end{figure}

This phenomenon aligns with challenges previously identified in the literature on fine-tuning generative models on their own outputs~\citep{shumailov2024ai,shumailov2023curse,alemohammad2024self}.  To illustrate this, we fine-tune Stable Diffusion with all advantages set to 1 to eliminate the reward signal. This is equivalent to fine-tuning on self-generation data in an online manner. We explore CFG scales of 2 and 4 in Figure~\ref{fig:fpo_image_supp}. We find that both CFG scales induce quality regression. Specifically, the CFG scale of 2 makes the generation more blurry, while the scale of 2 causes the generated images to feature high saturation and geometry patterns. Both eventually diverge to abstract geometric patterns.

\end{document}